%% file: arxiv_main.tex
\documentclass[11pt, a4paper]{lumia}

\usepackage[sort&compress]{natbib}
\input{notations}

\usepackage{graphicx}           
\usepackage{tikz}               
\usepackage[edges]{forest}      

\usepackage{url}                
\usepackage{xurl}               

\usepackage{array}              
\usepackage{longtable}          
\usepackage{multirow}           
\usepackage{makecell}           
\usepackage{ragged2e}           

\usepackage{mathtools}          
\usepackage{nicefrac}           

\usepackage{algorithm}          
\usepackage{algorithmicx}       
\usepackage{algpseudocode}      
\usepackage{listings}           

\usepackage{subcaption}         
\usepackage{wrapfig}            
\usepackage[export]{adjustbox}  

\usepackage{changes}            
\usepackage{xspace}             
\usepackage[normalem]{ulem}     
\usepackage{CJKutf8}            

\usepackage[tikz]{bclogo}       
\usepackage[framemethod=tikz]{mdframed} 

\usepackage{lipsum}             
\usepackage{tocloft}            
\usepackage{afterpage}          
\usepackage{bbding}             
\usepackage{epigraph}           
\usepackage{minitoc}            
\usepackage{multicol}           
\usepackage{textgreek}          

\newcolumntype{P}[1]{>{\RaggedRight\arraybackslash}p{#1}}

\definecolor{uclablue}{RGB}{39, 116, 174}
\definecolor{bigaired}{RGB}{156, 0, 0}
\definecolor{myblue}{HTML}{598BE7}
\definecolor{mildblue}{RGB}{31,119,180}
\definecolor{sectionblue}{RGB}{70, 130, 180}
\definecolor{methodblue}{RGB}{0, 150, 136}
\definecolor{bgblue}{RGB}{245,243,253}
\definecolor{ttblue}{RGB}{91,194,224}
\definecolor{mygreen}{rgb}{0.64, 0.56, 0.88}
\definecolor{myyellow}{rgb}{0.68, 0.6, 0.1}
\definecolor{fancygreen}{rgb}{0.33, 0.68, 0.20}
\definecolor{salmon}{rgb}{0.94, 0.52, 0.49}
\definecolor{tablegreen}{rgb}{0.82, 0.94, 0.75}
\definecolor{tableblue}{rgb}{0.81, 0.90, 0.94}
\definecolor{tablered}{rgb}{0.97, 0.85, 0.85}
\definecolor{tableorange}{rgb}{0.96, 0.85, 0.81}
\definecolor{myorange}{rgb}{1.0, 0.49, 0.0}
\definecolor{tlgreen}{rgb}{0.33, 0.68, 0.20}
\definecolor{darkgreen}{RGB}{0,100,0}
\definecolor{darkred}{RGB}{200, 0, 0}
\definecolor{customyellow}{HTML}{FFFACD}
\definecolor{refinegreen}{RGB}{0, 128, 75}
\definecolor{scoregreen}{RGB}{34, 139, 34}
\definecolor{hidden-blue}{RGB}{194,232,247}
\definecolor{hidden-black}{RGB}{20,68,106}
\definecolor{yes}{HTML}{C6EFCE}
\definecolor{no}{HTML}{FFC7CE}
\definecolor{partial}{HTML}{FFEB9C}
\definecolor{external}{HTML}{D9E1F2}
\definecolor{hdr}{HTML}{F2F2F2}
\definecolor{GRPOrow}{gray}{0.96}
\definecolor{FlowRLrow}{RGB}{225,236,255}
\definecolor{FlowBlue}{RGB}{80,120,210}
\definecolor{GRPOGray}{gray}{0.35}
\definecolor{mygrey}{RGB}{200,200,200}

\setlist[itemize]{leftmargin=20pt, noitemsep, topsep=0pt}

\newcommand{\cmark}{\textcolor{darkgreen}{\boldmath$\checkmark$}}
\newcommand{\xmark}{\textcolor{darkred}{\boldmath$\times$}}

\newenvironment{itemize*}%
 {\leftmargini=10pt\begin{itemize}%
  \setlength{\itemsep}{0pt}%
  \setlength{\parskip}{0pt}%
  }%
 {\end{itemize}}

\newenvironment{enumerate*}%
 {\begin{enumerate}%
  \setlength{\itemsep}{0pt}%
  \setlength{\parskip}{0pt}}%
 {\end{enumerate}}

\newcommand{\cellstatus}[1]{%
  \begingroup
  \StrTrim{#1}[\statusval]%
  \IfStrEq{\statusval}{Yes}{\cellcolor{yes}\cmark}{}%
  \IfStrEq{\statusval}{No}{\cellcolor{no}\xmark}{}%
  \IfBeginWith{\statusval}{Yes (}{\cellcolor{yes}\cmark~\textit{\statusval\unskip}}{}%
  \IfStrEq{\statusval}{Partial}{\cellcolor{partial}\textbf{Partial}}{}%
  \IfStrEq{\statusval}{External}{\cellcolor{external}\textbf{External}}{}%
  \endgroup
}

\newtcolorbox{myboxi}[1][]{
  breakable,
  title=#1,
  colback=red!5,
  colbacktitle=red!5,
  coltitle=black,
  fonttitle=\bfseries,
  bottomrule=0pt,
  toprule=0pt,
  leftrule=2pt,
  rightrule=2pt,
  titlerule=0pt,
  arc=0pt,
  outer arc=0pt,
  colframe=red,
}

\newtcolorbox{myboxnote}[1][]{
  breakable,
  title=#1,
  colback=orange!0,
  colbacktitle=orange!0,
  coltitle=black,
  fonttitle=\bfseries,
  bottomrule=0pt,
  toprule=0pt,
  leftrule=2pt,
  rightrule=2pt,
  titlerule=0pt,
  arc=0pt,
  outer arc=0pt,
  colframe=orange,
}

\newtcolorbox{myboxii}[1][]{
  breakable,
  freelance,
  title=#1,
  colback=white,
  colbacktitle=white,
  coltitle=black,
  fonttitle=\bfseries,
  bottomrule=0pt,
  boxrule=0pt,
  colframe=white,
  overlay unbroken and first={
  \draw[red!75!black,line width=3pt]
    ([xshift=5pt]frame.north west) -- 
    (frame.north west) -- 
    (frame.south west);
  \draw[red!75!black,line width=3pt]
    ([xshift=-5pt]frame.north east) -- 
    (frame.north east) -- 
    (frame.south east);
  },
  overlay unbroken app={
  \draw[red!75!black,line width=3pt,line cap=rect]
    (frame.south west) -- 
    ([xshift=5pt]frame.south west);
  \draw[red!75!black,line width=3pt,line cap=rect]
    (frame.south east) -- 
    ([xshift=-5pt]frame.south east);
  },
  overlay middle and last={
  \draw[red!75!black,line width=3pt]
    (frame.north west) -- 
    (frame.south west);
  \draw[red!75!black,line width=3pt]
    (frame.north east) -- 
    (frame.south east);
  },
  overlay last app={
  \draw[red!75!black,line width=3pt,line cap=rect]
    (frame.south west) --
    ([xshift=5pt]frame.south west);
  \draw[red!75!black,line width=3pt,line cap=rect]
    (frame.south east) --
    ([xshift=-5pt]frame.south east);
  },
}

\tcbset{
  takeawaysbox/.style={
    title=Takeaways,
    colback=lightblue!80,
    colframe=black,
    fonttitle=\bfseries\small,
    coltitle=white,
    colbacktitle=black,
    enhanced,
    attach boxed title to top left={xshift=2.5mm,yshift=-2.5mm},
    boxed title style={rounded corners, size=small, colframe=black, colback=black},
    width=\linewidth,
    arc=3.5mm
  }
}

\mdfdefinestyle{mystyle}{%
  rightline=true,
  innerleftmargin=10,
  innerrightmargin=10,
  outerlinewidth=3pt,
  topline=false,
  rightline=true,
  bottomline=false,
  skipabove=\topsep,
  skipbelow=\topsep
}

\tikzset{%
    every node/.style={font=\tiny},
    parent/.style =          {align=center,text width=2cm,rounded corners=3pt, line width=0.3mm, fill=gray!10,draw=gray!80},
    child/.style =           {align=center,text width=2.0cm,rounded corners=3pt, fill=blue!10,draw=blue!80,line width=0.3mm},
    grandchild/.style =      {align=center,text width=2cm,rounded corners=3pt},
    greatgrandchild/.style = {align=center,text width=1.5cm,rounded corners=3pt},
    greatgrandchild2/.style = {align=center,text width=1.5cm,rounded corners=3pt},    
    referenceblock/.style =  {align=center,text width=1.5cm,rounded corners=2pt},
    pretrain/.style =           {align=center,text width=2.0cm,rounded corners=3pt, fill=blue!10,draw=blue!80,line width=0.3mm},   
    pretrain_work/.style =           {align=center, text width=8.5cm,rounded corners=3pt, fill=blue!10,draw=blue!0,line width=0.3mm},  
    template/.style =           {align=center,text width=2.0cm,rounded corners=3pt, fill=red!10,draw=red!80,line width=0.3mm},   
    template_work/.style =           {align=center,text width=8.5cm,rounded corners=3pt, fill=red!10,draw=red!0,line width=0.3mm},    
    answer/.style =           {align=center,text width=2.0cm,rounded corners=3pt, fill= cyan!10,draw= cyan!80,line width=0.3mm},   
    answer_work/.style =           {align=center,text width=8.5cm,rounded corners=3pt, fill= cyan!10,draw= cyan!0,line width=0.3mm},      
    multiple/.style =           {align=center,text width=2.0cm,rounded corners=3pt, fill= orange!10,draw= orange!80,line width=0.3mm},   
    multiple_work/.style =           {align=center,text width=8.5cm,rounded corners=3pt, fill= orange!10,draw= orange!0,line width=0.3mm},        
    tuning/.style =           {align=center,text width=2.0cm,rounded corners=3pt, fill= magenta!10,draw= magenta!80,line width=0.3mm},   
    tuning_work/.style =           {align=center,text width=8.5cm,rounded corners=3pt, fill= magenta!10,draw= magenta!0,line width=0.3mm},          
}

\newcommand{\lstbg}[3][0pt]{{\fboxsep#1\colorbox{#2}{\strut #3}}}

\lstdefinelanguage{diff}{
  basicstyle=\ttfamily\small,
  morecomment=[f][\lstbg{red!20}]-,
  morecomment=[f][\lstbg{green!20}]+,
}

\lstdefinelanguage{diffpython}{
  language=diff,
  morekeywords={def, if, else, for, while, return, import, from, as, class, with, try, except, finally, raise, lambda, and, or, not, in, is, None, True, False},
  morecomment=[l]{\#},
  morestring=[b]",
  morestring=[b]',
}

\usepackage{xspace}

\usepackage{xcolor}

\definecolor{ForestGreen}{RGB}{34,139,34}
\definecolor{myyellow}{RGB}{181, 181, 27}

\usepackage{amsmath}
\usepackage{amssymb}
\usepackage{mathtools}
\usepackage{amsthm}
\usepackage{fontawesome5} 

\usepackage[capitalize,noabbrev]{cleveref}

\usepackage{multirow}
\usepackage{makecell}
\usepackage{enumerate}
\usepackage{enumitem}
\usepackage{pifont}

\usepackage[ruled,algo2e,vlined]{algorithm2e}
\SetKwInOut{Input}{Input}\SetKwInOut{Output}{Output}
\SetKwComment{Comment}{$\triangleright$\ }{}

\definecolor{darkgreen}{RGB}{30, 130, 30}
\definecolor{cream}{RGB}{253, 250, 242}
\renewcommand{\cmark}{\textcolor{darkgreen}{\ding{51}}} 
\renewcommand{\xmark}{\textcolor{red}{\ding{55}}}       

\usepackage{listings}
\usepackage{lineno}

\input{preamble}

\usepackage{ulem}
\usepackage{pifont}

\setheadertext{LUMIA Lab}

\githublink{https://github.com/neu-3dvi/MESSI/} 
\websitelink{https://neu-3dvi.github.io/MESSI/}

\title{MESSI: Multi-Expert Soccer Skill Integration for Physics-Based Character Control}

\author{%
 Xiaogang Peng$^{1*}$, Jiaqi Zou$^{2*}$, Lei Zhong$^3$, Sirui Xu$^4$, Huaizu Jiang$^{1}$\\
  $^1$Northeastern University \hspace{1mm}
  $^2$Duke University \hspace{1mm}
  $^3$University of Edinburgh \hspace{1mm}
  $^4$University of Illinois at Urbana-Champaign
  \\
}

\begin{document}

\input{sec_arxiv/0_abstract}  

\maketitle

\begin{figure*}[h]
\centering
\includegraphics[width=1.0\textwidth]{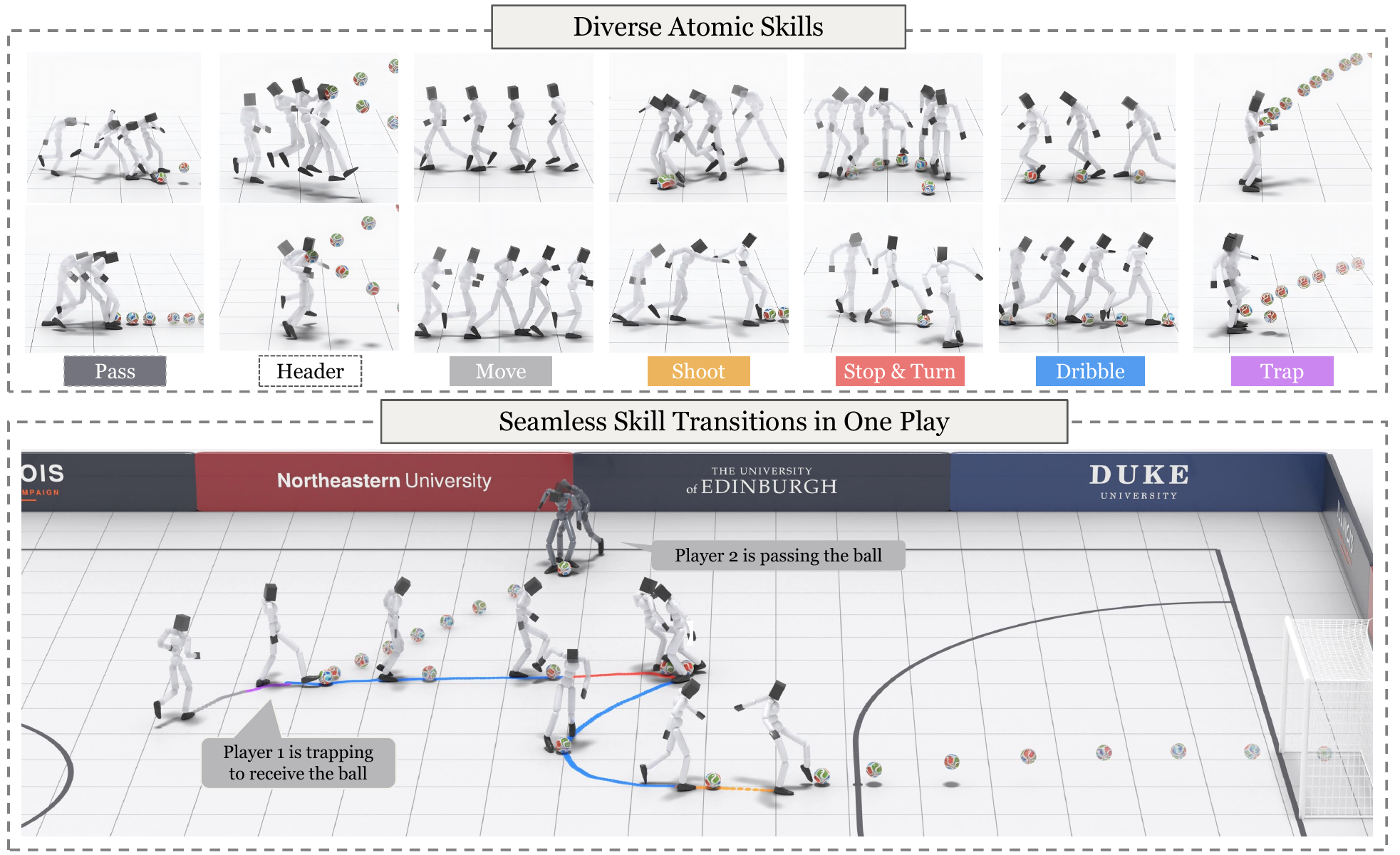}
\captionsetup{hypcap=false}
\vspace{-2em}
\caption{\textbf{MESSI: Composable physics-based soccer control.}
Our unified controller integrates eight soccer skills
and enables smooth transitions between them.
\textbf{Top:} Representative atomic skills.
\textbf{Bottom:} A continuous play combining
passing, trapping, dribbling, and shooting.}
\captionsetup{hypcap=true}
\label{fig:teaser}%
\end{figure*}

\input{sec_arxiv/1_intro}

\input{sec_arxiv/2_related}

\input{sec_arxiv/3_method}

\input{sec_arxiv/4_experiments}
\input{sec_arxiv/5_conclusion}

\clearpage

{
    \small
    \bibliography{references}
}

\input{sec_arxiv/X_suppl}

\end{document}

%% file: notations.tex
\usepackage{xspace}

\newcommand*{\ie}{\textit{i.e.}\xspace}

\theoremstyle{plain}

\newtheorem*{proposition*}{Proposition}

\theoremstyle{definition}

\theoremstyle{definition}

\def\Figref#1{Figure~\ref{#1}}

\def\Secref#1{Section~\ref{#1}}

\def\eqref#1{equation~\ref{#1}}
\def\Eqref#1{Equation~\ref{#1}}



%% file: preamble.tex
\usepackage{cuted} 

\usepackage{currfile} 

\usepackage{caption} 

\newcommand{\paper}{MESSI\xspace}

\definecolor{bestcell}{HTML}{DCEFD9}
\newcommand{\best}[1]{\cellcolor{bestcell}\textbf{#1}}
\definecolor{secondcell}{HTML}{FCE4C5}
\newcommand{\second}[1]{\cellcolor{secondcell}#1}

\def\Figref#1{Figure~\ref{#1}}

\def\Secref#1{Section~\ref{#1}}

\def\eqref#1{equation~\ref{#1}}
\def\Eqref#1{Equation~\ref{#1}}

\definecolor{skMove}{HTML}{A8A8AE}
\definecolor{skDribble}{HTML}{5B9BD5}
\definecolor{skShoot}{HTML}{F0B050}
\definecolor{skPass}{HTML}{6B6B73}
\definecolor{skHeader}{HTML}{8C8C94}
\definecolor{skTrap}{HTML}{C58BE0}
\definecolor{skStop}{HTML}{E57373}
\definecolor{skGetup}{HTML}{7FB77E}
\newcommand{\skilltag}[2]{{\setlength{\fboxsep}{1pt}\colorbox{#1}{\textcolor{white}{\scriptsize\textsf{\textbf{#2}}}}}}
\newcommand{\Mv}{\skilltag{skMove}{Mv}}
\newcommand{\Dr}{\skilltag{skDribble}{Dr}}
\newcommand{\Sh}{\skilltag{skShoot}{Sh}}
\newcommand{\Pa}{\skilltag{skPass}{Pa}}
\newcommand{\Hd}{\skilltag{skHeader}{Hd}}
\newcommand{\Tr}{\skilltag{skTrap}{Tr}}
\newcommand{\ST}{\skilltag{skStop}{ST}}
\newcommand{\Gu}{\skilltag{skGetup}{Gu}}
\newcommand{\nxt}{\,{\scriptsize$\rightarrow$}\,}
\providecommand{\best}[1]{\cellcolor{green!15}\textbf{#1}}
\providecommand{\second}[1]{\cellcolor{orange!20}#1}

\definecolor{appblue}{RGB}{50,85,120}

\newcommand{\appsec}[2]{%
    \par\addvspace{0.35em}
    \noindent
    \hyperref[#1]{%
        \textbf{\textcolor{appblue}{\ref*{#1}\enspace #2}}%
    }%
    \nobreak\dotfill\nobreak
    \hyperref[#1]{\pageref*{#1}}%
    \par
}

\newcommand{\appsub}[2]{%
    \noindent\hspace*{1.2em}%
    \hyperref[#1]{%
        \ref*{#1}\enspace #2%
    }%
    \nobreak\dotfill\nobreak
    \hyperref[#1]{\pageref*{#1}}%
    \par
}

%% file: sec_arxiv/0_abstract.tex
\begin{abstract}

Physics-based soccer characters must perform diverse motor skills and transition smoothly between them, such as dribbling into a shot. However, skills that perform well in isolation may fail when executed sequentially, as preceding behaviors can leave the character and ball in unfamiliar states. We present \textbf{MESSI} (\textbf{M}ulti-\textbf{E}xpert \textbf{S}occer \textbf{S}kill \textbf{I}ntegration), a framework that enables multiple soccer skill transitions through shared adaptation while preserving the original skill policy outside transitions. MESSI first trains eight task-competent experts using motion priors and task rewards, then distills them into a unified, skill-conditioned Transformer. Rather than retraining individual experts for different transitions, we train a shared, context-conditioned low-rank adapter on the frozen student through online reinforcement learning. Activated upon skill switching, the adapter helps the target skill handle unfamiliar entry states, while a learned exit mechanism determines when to return control to the original student. By disabling the adapter during native skill execution, MESSI preserves the original student policy by construction. Experiments across individual skills, skill transitions, and continuous skill chains demonstrate substantial improvements over direct switching, including an increase in dribbling-to-shooting success from 19.8\% to 61.8\%.

\end{abstract}

%% file: sec_arxiv/1_intro.tex
\section{Introduction}
\label{sec:introduction}

\begin{wrapfigure}{r}{0.5\textwidth}
\centering
\includegraphics[width=1.0\linewidth]{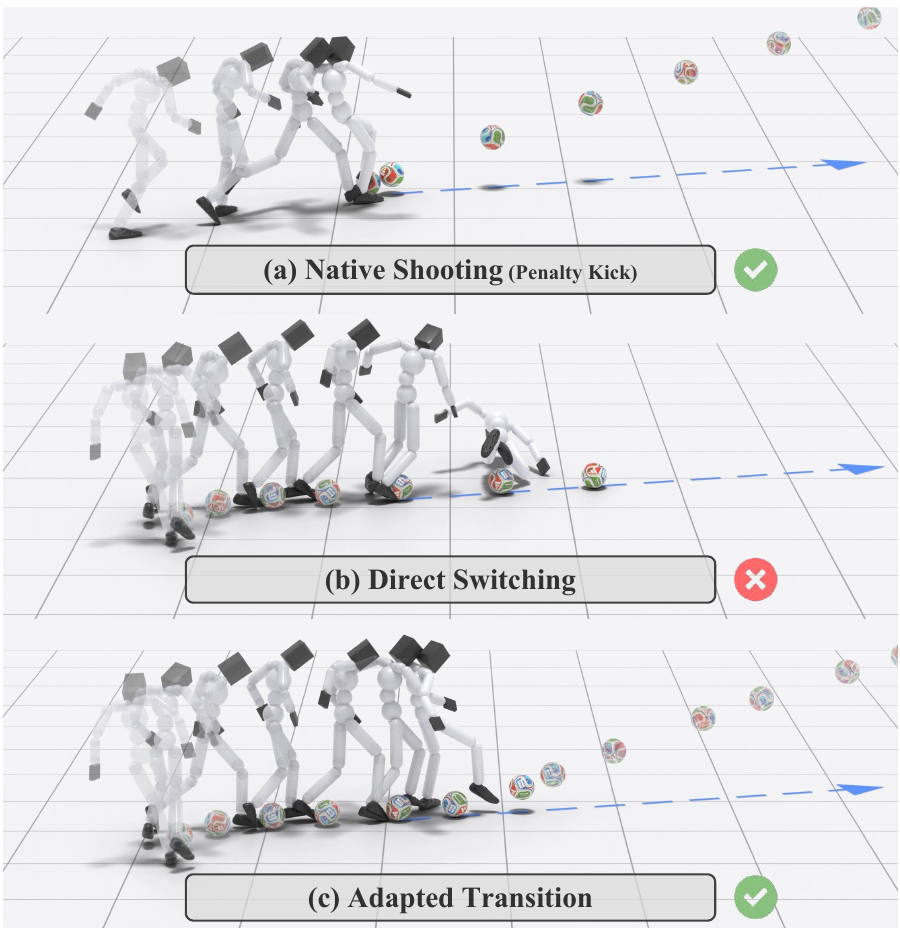}
\caption{\textbf{Challenges of skill transitions in physics-based soccer control.} (a) Shooting succeeds under native initialization. (b) Direct switching from dribbling to shooting fails due to unfavorable body and ball configurations. (c) With our adapter, active from the switch, the same handover leads to a successful shot. Overlaid poses illustrate motion sequences, and dashed arrows indicate the target shooting direction.}
\label{fig:transition_challenge}
\end{wrapfigure}

Soccer presents a challenging setting for physics-based character control, combining whole-body coordination with precise interactions with a dynamically moving ball~\citep{hong2019soccer,liu2022football,
haarnoja2024soccer,kim2025physicsfc,ye2026skillx}. A versatile soccer character must perform diverse skills, including locomotion, dribbling, shooting, passing, and trapping, while transitioning between them at speed. For example, a player may shoot immediately after dribbling or receive a pass while running. Such transitions are challenging because the character's posture, momentum, and ball configuration depend on the preceding behavior and may differ substantially from the states encountered during the target skill's training. Consequently, even well-trained individual skills can fail when executed sequentially, as illustrated in~\Figref{fig:transition_challenge}.

Recent advances in physics-based character control have enabled reusable motor skills through motion tracking and large-scale motion pretraining~\citep{peng2018deepmimic,peng2022ase,
tessler2024maskedmimic,juravsky2024superpadl,
shi2026gpc}. However, tracking reference motions alone does not guarantee task-specific ball control or reliable transitions absent from the motion data. While independently trained, task-competent experts can optimize individual skill objectives, composing their behaviors remains challenging when the target skill encounters unfamiliar body and ball states.
Existing approaches address skill transitions in several ways. Some learn skills and transitions jointly, either through unified multi-skill policies~\citep{ye2026skillx} or augmented interaction trajectories~\citep{yu2025skillmimic}. Such approaches may require revisiting joint policy training or transition data when introducing new transitions, potentially affecting previously learned behaviors. Others employ transition-aware initialization or fine-tuning using predecessor-generated states~\citep{kim2025physicsfc,narula2026joga}, coupling transition adaptation to individual skill policies. For more challenging transitions, policy composition methods additionally introduce intermediate behaviors when necessary~\citep{xu2025learning}, which can increase transition-specific training and maintenance overhead. While these approaches demonstrate effective skill composition, adapting a growing repertoire of skills without repeatedly modifying individual experts or introducing separate transition modules remains challenging. This motivates a shared adaptation strategy that supports diverse skill transitions while limiting interference with established skill behaviors.

To this end, we present \textbf{MESSI} (\textbf{M}ulti-\textbf{E}xpert \textbf{S}occer \textbf{S}kill \textbf{I}ntegration), a framework for integrating task-competent soccer skills and enabling their transitions within a unified physics-based controller. MESSI incorporates eight skills: off-ball movement, dribbling, shooting, passing, heading, trapping, stop-and-turn, and getting up. We first train skill-specific experts using motion priors, task rewards, and tailored curricula. Their behaviors are then distilled into a shared skill-conditioned student through a unified observation space.
Consolidating the experts into a single skill-conditioned policy provides a common backbone for transition adaptation. This enables different skill pairs to share a single adaptation module, rather than requiring separate adaptations of individual expert policies.


Our key insight is that transition adaptation is needed only after a skill switch, where the target skill encounters unfamiliar states generated by its predecessor.
Rather than modifying the target skill everywhere, we introduce a shared, gated, context-conditioned low-rank adapter on the frozen student. The gate opens when the
skill switches, and a learned exit decides when to close it, allowing the frozen student to resume control. Outside transitions, the adapter
never acts. We train the adapter through online reinforcement learning on predecessor-generated states and target-skill rewards. Conditioning the adapter on the skill pair and target command allows a single adaptation module to serve multiple transitions. New transition pairs can be incorporated into the shared adaptation process without modifying the original experts or the student backbone, avoiding the need to maintain separate transition-specific modules. This design improves transition robustness while leaving the native skills exactly unchanged. 

Our main contributions are summarized as follows:
\begin{itemize}
    \item We develop a physics-based soccer control framework that consolidates eight independently trained, task-competent experts into a unified skill-conditioned policy, providing a shared backbone for transition adaptation.
    \item We propose an online, shared transition adaptation method using a context-conditioned low-rank adapter, enabling joint adaptation across multiple skill pairs without retraining individual experts.
    \item We introduce a learned exit mechanism that dynamically determines the duration of transition adaptation, allowing the controller to return to the frozen student when adaptation is no longer needed, while preserving the original policy during native skill execution.
\end{itemize}

%% file: sec_arxiv/2_related.tex
\section{Related Work}
\label{sec:related_work}
\paragraph{Physics-based interaction skills.}
Physics-based character control has progressed from motion
tracking~\citep{peng2018deepmimic} and adversarial motion
priors~\citep{peng2021amp,peng2022ase} to scalable,
unified controllers~\citep{tessler2024maskedmimic,shi2026gpc}.
Athletic ball interactions introduce additional challenges
of coordinating whole-body motion with dynamic objects
and precisely timed contacts. Early work learned basketball
dribbling through trajectory optimization and reinforcement
learning~\citep{liu2018basketball}, and synthesized soccer
dribbling and shooting through coupled character--ball
control~\citep{hong2019soccer}.
Recent methods have demonstrated basketball interaction
skills~\citep{wang2025skillmimic}, agile humanoid
soccer~\citep{haarnoja2024soccer}, accurate
shooting~\citep{zhong2026robonaldo}, and vision-based
soccer control~\citep{jin2026davis}.
These advances motivate learning task-competent skills
that optimize physical interaction objectives beyond
reference-motion reproduction.

\paragraph{Skill composition and transitions.}
Prior work addresses skill composition through several
strategies. Some methods learn skills and transitions
jointly. ~\citep{xie2022soccerjuggling} employ
a layer-wise mixture-of-experts for soccer juggling and
skill switching, while SkillMimic-V2~\citep{yu2025skillmimic}
constructs augmented interaction trajectories and state
transition fields.
SkillX learns multiple soccer skills
in a unified policy~\citep{ye2026skillx} , and SoccerBot
combines latent motion priors with transition-state
initialization~\citep{shao2026soccerbot}.
Other methods adapt target skills to states generated by
their predecessors, including PhysicsFC~\citep{kim2025physicsfc}
and Joga~\citep{narula2026joga}.
Learning to Ball further combines
policy adaptation with intermediate policies for challenging
transitions~\citep{xu2025learning}.
While effective, extending these approaches to more
skills and transitions can require updating shared policies,
augmenting transition data, or maintaining specialized
adaptation procedures.
We instead study transition adaptation after independently
trained skills have been consolidated into a single controller.

\paragraph{Expert distillation and policy adaptation.}
Policy distillation~\citep{rusu2016policydistillation} and
DAgger~\citep{ross2011dagger} provide foundations for
consolidating expert behaviors.
SuperPADL progressively
distills specialized RL experts into a scalable controller~\citep{juravsky2024superpadl},
while InterMimic combines
teacher--student learning with RL refinement for diverse
human-object interactions~\citep{xu2025intermimic}.
MaskedManipulator distills a motion-tracking controller into a goal-conditioned policy~\citep{tessler2025maskedmanipulator}, and TokenHSI uses task
tokenization for unified interaction control~\citep{pan2025tokenhsi}.
Post-training adaptation has also been explored through
policy refinement~\citep{xu2023adaptnet},
conditional low-rank updates~\citep{shi2026gpc},
object-aware residual learning~\citep{zhao2025resmimic},
and RL fine-tuning of distilled interaction
priors~\citep{xu2026interprior}.
In contrast, MESSI focuses on adapting multiple skill
transitions through a single context-conditioned adapter
on a frozen student, with a learned exit that restricts
adaptation to skill boundaries and preserves native-policy
behavior.

%% file: sec_arxiv/3_method.tex
\section{Method}
\label{sec:method}



As shown in~\Figref{fig:pipeline}, \paper consists of three stages: skill-specific expert learning, multi-skill distillation, and unified transition adaptation. At test time, the controller receives the current character and ball states, skill label, and task command. A user or high-level planner specifies the next skill and its command at a scheduled skill switch, without requiring reference motions.

\subsection{Skill-specific Expert Learning}
\label{sec:experts}

\begin{figure*}[t]
    \centering
    \includegraphics[width=1.0\linewidth]{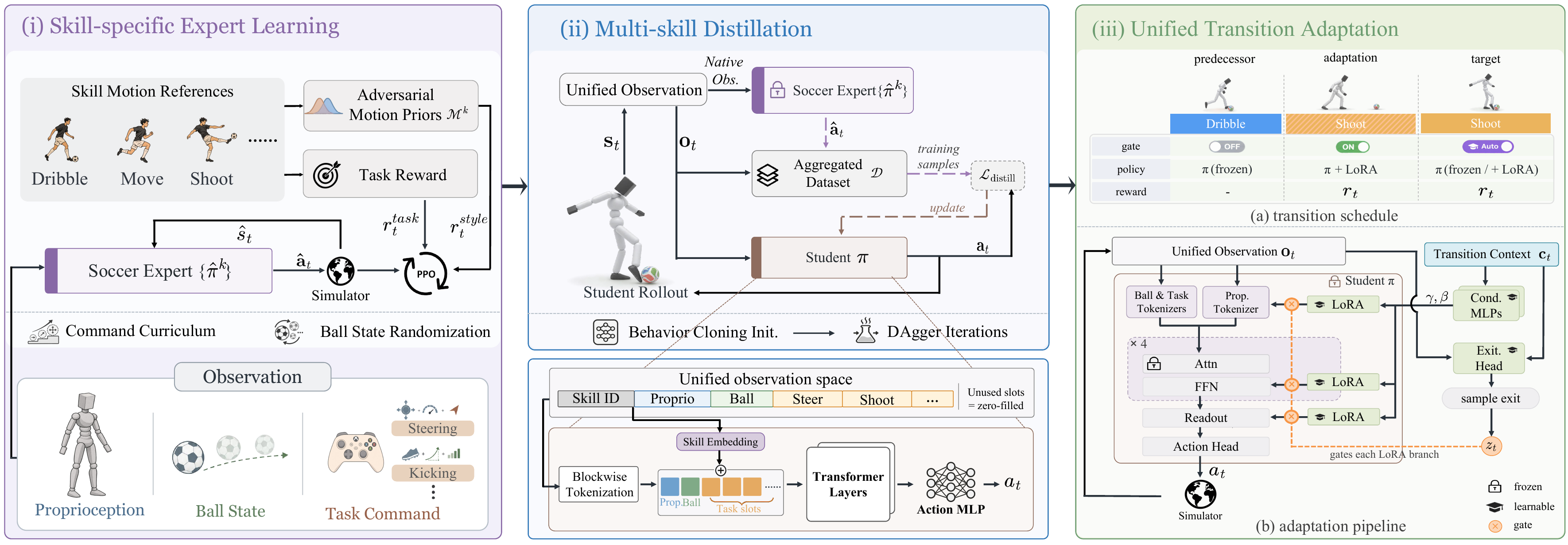}
    \caption{\textbf{Overview of MESSI.} (i)~\textit{Skill-specific expert learning:} each of the eight skills is trained with a motion prior and its own task reward.
    (ii)~\textit{Multi-skill distillation:} the experts are distilled into one
    skill-conditioned student over a unified observation space, by behavior cloning
    followed by DAgger. (iii)~\textit{Unified transition adaptation:} a shared conditional
    LoRA adapter on the frozen student acts only after a skill switch and is trained online
    on predecessor-generated states.}
    \label{fig:pipeline}
\end{figure*}

As shown in \Figref{fig:pipeline}~(i), we first train each expert policy $\hat{\pi}^k(\hat{\mathbf{a}}_t \mid \hat{\mathbf{s}}_t, \mathbf{g}^k_t)$
for each skill $k$. The state $\hat{\mathbf{s}}_t$ contains the character's proprioception and the ball state relative to the character, and is the same for all skills. The task command $\mathbf{g}^k_t$ is specific to the skill $k$. Actions are target joint angles for PD controllers. Each expert is trained in its own environment, with its own motion data, initial-state distribution, and termination rules. These choices are made per skill and are not constrained by the later integration stages.

\paragraph{Rewards and optimization.}
Each expert maximizes several objectives $j$: a task reward $r^{\text{task}}$
defined by the skill, and a style reward from the corresponding adversarial motion
prior $\mathcal{M}^{k}$~\citep{peng2021amp},
\begin{equation}
r^{\text{style}}_t = -\log\bigl(1 - D(\mathbf{o}_{t-H+1:t})\bigr),
\end{equation}
where $D$ is a discriminator over windows of $H$ character states. For the
kicking and heading skills, $D$ is also conditioned on the technique label, and
these skills add a small leg-pose objective that keeps the striking leg close
to the reference motion. Summing the objectives into one reward lets their
effective weights change with the reward scales during training. Inspired by~\citep{xu2023composite}, rather than combining task and style rewards into a single scalar objective, we optimize separately normalized advantages to maintain their relative contributions during training.


\paragraph{Shooting.}
The shooting command specifies the target launch velocity of the ball, the
striking foot, and the technique (instep or inside shot). The launch velocity is
recorded when the ball is struck. From then until the end of the episode, the task
reward pays $\exp(-k\lVert\mathbf{v}_{\text{launch}} - \mathbf{v}^*\rVert^2)$, with
a bonus if the commanded foot made the strike. The policy thus receives a dense
signal for the quality of a single, brief contact. Before the strike, an approach
term draws the striking foot to the ball. After the strike, a recovery term
rewards a balanced stance. Commands are sampled from technique-specific ranges
of speed, elevation, and direction, and the ball starts either at rest or rolling.

\paragraph{Dribbling.}
The dribbling command specifies a target direction and speed. The task reward
combines velocity tracking with a lead term that keeps the ball at a fixed
distance ahead of the character along the target direction. Lateral deviation of
the ball is penalized more strongly than longitudinal deviation. The
velocity-tracking reward is multiplied by a gate on the character--ball
distance. The gate is constant while the ball is within reach and decays beyond it,
so the policy gains nothing by running ahead and leaving the ball behind.

\paragraph{Other skills.}
The remaining skills follow the same template with the commands and task objectives, and
their details about rewards, command distributions, initialization, and termination of all eight skills are included in Appendix.

\subsection{Multi-Skill Distillation}
\label{sec:distill}

The experts differ in their commands, network inputs, and environments. We distill
them into a single student policy $\pi$ that executes every skill from one
observation space and selects the skill by a label.

\paragraph{Unified observation space.}
The student observes
\begin{equation}
\mathbf{o}_t = \bigl[\mathbf{s}_t,\; \mathbf{b}^1_t, \dots, \mathbf{b}^M_t,\; \mathbf{e}_{k_t}\bigr],
\label{eq:union}
\end{equation}
where $\mathbf{s}_t$ is the shared state of Sec.~\ref{sec:experts},
$\mathbf{b}^1_t, \dots, \mathbf{b}^M_t$ are task blocks, and $\mathbf{e}_{k_t}$ is
the one-hot label of the active skill $k_t$. The command of each skill is written
into one task block. Skills with the same kind of command share a block: dribbling
and off-ball movement, for example, both use the steering block. The blocks of the
other skills are set to zero. A fixed selection $P_k$ recovers the native observation of expert
$k$, $P_k\mathbf{o}_t = (\mathbf{s}_t, \mathbf{g}^k_t)$, so the expert of the
active skill can label every state the student visits.

\paragraph{Student architecture.}
The student is
a Transformer~\citep{vaswani2017attention} over the observation blocks. A proprioception tokenizer, a ball
tokenizer, and one task tokenizer per task block each map their block to a token,
and an embedding of the skill label is added to every token. Attention is masked:
the proprioception token attends to all tokens, while every other token attends
only to itself and the proprioception token. A task block shared by several
skills, such as steering, therefore never attends directly to a block that only
some of them use, such as the ball. A readout layer concatenates the output
tokens, and an action layer maps the result to the mean action
$\boldsymbol{\mu}(\mathbf{o}_t)$.

\paragraph{Distillation.}
The student is trained to match the experts' mean actions on an aggregated dataset
$\mathcal{D}$ of observations and skill labels,
\begin{equation}
\small{
\mathcal{L}_{\text{distill}} = \mathbb{E}_{(\mathbf{o}, k)\sim\mathcal{D}}
\Bigl[\bigl\lVert \boldsymbol{\mu}(\mathbf{o}) - \hat{\boldsymbol{\mu}}_k(P_k\,\mathbf{o})\bigr\rVert^2\Bigr],
}
\label{eq:distill}
\end{equation}
where $\boldsymbol{\hat{\mu}}_k$ is the mean action of expert $\hat{\pi}^k$. We first fill
$\mathcal{D}$ with rollouts of each expert in its own environment, with the
initial states, commands, and terminations of its training, and initialize the
student by behavior cloning. The student's errors then lead it to states that the
experts rarely visit. We therefore apply DAgger~\citep{ross2011dagger}. In each
round, the student executes its mean action in the environment of every skill, the
expert of that skill labels each visited state, the new samples are added to
$\mathcal{D}$, and the student is retrained on all of $\mathcal{D}$. Every skill
contributes the same number of samples, and the experts stay frozen.
Appendix provides the architecture and the training schedule.

Each rollout executes a single skill from that skill's own initial states.
Distillation therefore never trains the student to start a skill from the state
that its predecessor leaves behind. Sec.~\ref{sec:adapter} addresses these
transitions.

\subsection{Unified Transition Adaptation}
\label{sec:adapter}

Although the distilled student preserves individual skill capabilities, directly switching skills can expose the target policy to unfamiliar body and ball states. Fine-tuning on these states may improve transitions but alter the target skill's performance under native initialization. Our approach instead adapts the frozen student through a single shared module, with two key design choices: conditioning adaptation on the skill transition and learning when to deactivate it.

\paragraph{Transition schedule.}

As illustrated in \Figref{fig:pipeline}(iii), the student executes the predecessor skill with the adapter disabled ($z_t=0$). At a skill switch, the target skill becomes active and the adapter is enabled ($z_t=1$) to handle the unfamiliar entry state. A learned exit policy subsequently determines when to deactivate the adapter, returning control to the original student. Native skill execution always bypasses the adapter.

\paragraph{Shared conditional adapter.}
Rather than training separate policies or adapters for each skill pair, we use a shared low-rank adapter conditioned on the transition context $c_t$, which encodes the predecessor and target skills, target command, and elapsed time since the switch. Following conditional low-rank adaptation~\citep{shi2026gpc, hu2021lora}, each adapted linear layer computes
\begin{equation}
y = W_0x +
z_t B\left[(Ax)\odot(1+\gamma(c_t))+\beta(c_t)\right],
\label{eq:cond_lora}
\end{equation}
where $W_0$ is frozen, $A$ and $B$ are trainable low-rank matrices, and $\gamma(c_t)$ and $\beta(c_t)$ modulate the update according to the transition context. This allows the same adapter parameters to serve multiple skill pairs, while $z_t=0$ recovers the original student. We apply it to the proprioception tokenizer, Transformer feed-forward layers, and readout.

\paragraph{Learned exit policy.}
The required adaptation duration varies across transitions and encountered states. A fixed window may terminate adaptation prematurely or keep the adapter active longer than necessary.
We therefore learn a state-dependent exit policy that predicts the probability of deactivating the adapter from the observation $o_t$ and transition context $c_t$.
The exit policy is trained jointly with the adapter using PPO, balancing target-task rewards against a per-step adapter usage cost.
Rewards obtained after an exit also contribute to the return, so the exit decision accounts for the subsequent execution of the frozen student. Exit decisions are allowed after $0.2$\,s, and the first sampled exit closes the gate until the next skill switch.

\paragraph{Training.}
After a skill switch, we optimize the target-skill reward with an adapter usage cost and failure penalty:
\begin{equation}
r_t = r_l^{\mathrm{task}}(s_t) - 0.02z_t - p_t,
\end{equation}
where $p_t$ is $1$ for a fall or ball loss and $0.5$ for other failures.
The adapter and exit policy are optimized jointly using PPO, with the importance ratio computed from the joint log-probability of actions and exit decisions. Only steps with an active adapter contribute policy gradients. The critic is conditioned on $c_t$. All transitions are trained jointly through online simulation with shared adapter parameters.

\input{tables_arxiv/table_macros}
\input{tables_arxiv/atomic_skills}

\input{tables_arxiv/transition_main}

%% file: tables_arxiv/table_macros.tex
\providecommand{\messipending}{\textit{TBD}}
\providecommand{\messina}{\textemdash}

%% file: tables_arxiv/atomic_skills.tex
\begin{table*}[t]
  \centering
  \setlength{\tabcolsep}{4pt}
  \renewcommand{\arraystretch}{1.2}

   \caption{\textbf{Atomic skills and distillation fidelity.} 1024 paired native cases per skill. Rates in \%, velocity errors in m/s. The student runs with the adapter off. Green: best; orange: second best.}
 \vspace{-0.5em}

  \label{tab:atomic_skills}
  \resizebox{\textwidth}{!}{%
  \begin{tabular}{lccccccccccc}
    \toprule
    & \multicolumn{3}{c}{Off-ball}
    & \multicolumn{2}{c}{Dribble}
    & Shoot & Pass & Header & Trap & Stop\&Turn & Getup \\
    \cmidrule(lr){2-4}\cmidrule(lr){5-6}
    \cmidrule(lr){7-7}\cmidrule(lr){8-8}\cmidrule(lr){9-9}
    \cmidrule(lr){10-10}\cmidrule(lr){11-11}\cmidrule(lr){12-12}
    Method
    & \shortstack{Success\,(\%) $\uparrow$}
    & \shortstack{Velocity error\\(m/s) $\downarrow$}
    & \shortstack{Fail\,(\%) $\downarrow$}
    & \shortstack{Velocity error\\(m/s) $\downarrow$}
    & \shortstack{Fail\,(\%) $\downarrow$}
    & \shortstack{Bearing\\(\%) $\uparrow$}
    & \shortstack{Bearing\\(\%) $\uparrow$}
    & \shortstack{Legal touch\\+ bearing (\%) $\uparrow$}
    & \shortstack{Success\\(\%) $\uparrow$}
    & \shortstack{Trap latch\\(\%) $\uparrow$}
    & \shortstack{Recovery\\(\%) $\uparrow$} \\
    \midrule

    TokenHSI*~\citep{pan2025tokenhsi}
      & 34.47 & 1.10 & 15.04
      & 2.830 & \best{11.43}
      & 2.73 & 7.03 & 3.12
      & 23.63 & 43.85 & 95.02 \\

     SkillX*~\citep{ye2026skillx}
      & 56.48 & 0.726 & 5.79
      & 2.324 & 19.31
      & 3.91 & 9.96 & 5.18
      & 30.76 & 72.07 & 94.63 \\

      Expert only
      & \best{69.43} & \best{0.617} & \best{1.76}
      & \best{1.989} & \second{12.60}
      & \best{79.88} & \best{67.58} & \second{22.46}
      & \best{97.46} & \best{91.70} & \best{99.12} \\
     BC student only
      & 29.98 & 0.919 & 43.16 & 2.281 & 28.03
      & 59.77 & 55.08 & 22.36 & 70.21 & 88.38 & 98.73 \\
    BC+DAgger student (Ours)
      & \second{68.95} & \second{0.639} & \second{3.12} & \second{2.104} & {17.77}
      & \second{77.93} & \second{63.77} & \best{22.66} & \second{95.02} & \second{91.21} & \second{99.02} \\
    \bottomrule
  \end{tabular}%
  }
  \vspace{-0.5em}
\end{table*}

%% file: tables_arxiv/transition_main.tex
\begin{table}[h]
  \centering
    \renewcommand{\arraystretch}{1.0}
  \caption{\textbf{Skill transition performance and native-policy preservation.} Transition success rates (\%) from identical predecessor-generated states, alongside native-skill performance.}
  \vspace{-0.5em}
  \label{tab:transitions}
  \resizebox{0.8\linewidth}{!}{%
  \begin{tabular}{cccccc|cc}
    \toprule
    & \multicolumn{5}{c|}{Transition success (\%) $\uparrow$}
    & \multicolumn{2}{c}{ {Native-policy}} \\
    \cmidrule(lr){2-6}\cmidrule(lr){7-8}
    Method
    & \shortstack{Dribble\\$\to$Shoot}
    & \shortstack{Stop\\$\to$Dribble}
    & \shortstack{Trap\\$\to$Dribble}
    & \shortstack{Getup\\$\to$Dribble}
    & \shortstack{Off-ball\\$\to$Dribble}
    & \shortstack{Shoot\\(\%) $\uparrow$}
    & \shortstack{Dribble vel.\\error (m/s) $\downarrow$} \\
    \midrule

     SkillX*~\citep{ye2026skillx}
      & 2.7 & {95.8} & {47.9} & \second{91.3} & {83.9}
      & - & - \\ \midrule
    
     \shortstack{Student\\ \small{(direct switch)}}
      & 19.8 & 44.1 & 20.7 & 32.6 & 62.7
      & 77.93 & 2.104  \\    \midrule
    \shortstack{Expert\\ \small{+ STI}}
      & {54.1} & 63.8 & 27.7 & 55.7 & 59.8
      & 63.09 & 2.662  \\    \midrule
 Full FT
      & {57.3} & \second{96.9} & \second{49.1} & {91.0} & \second{95.0}
      & 54.82 & {2.152} \\ \midrule
     Per-trans.
      & \best{67.8} & {94.0} & {47.8} & {90.8} & {92.6}
      & 77.93 & 2.104 \\ \midrule
    
    Ours
      & \second{61.8} & \best{97.0} & \best{50.4} & \best{91.7} & \best{96.3}
      & 77.93 & 2.104 \\
    \bottomrule
  \end{tabular}%
  }
\end{table}

%% file: sec_arxiv/4_experiments.tex
\section{Experiments}

We evaluate MESSI at three levels: single skills from their native initial states (Sec.~\ref{sec:exp_atomic}), single transitions from states produced by a predecessor skill (Secs.~\ref{sec:exp_transitions} and~\ref{sec:exp_exit}), and continuous chains of skills (Sec.~\ref{sec:exp_chains}). GPU PhysX rollouts differ across machines, so every paired comparison runs in the same machine from identical initial states.

\subsection{Implementation Details}
\label{sec:exp_impl}

All skills are simulated in Isaac Lab~\citep{mittal2023orbit} with GPU PhysX at
240\,Hz, and all policies run at 60\,Hz, the frame rate of the reference motions.
The character is a 28-DoF humanoid driven by PD controllers, with shoe-shaped collision meshes for the feet. The rigid-body solver models neither aerodynamics nor rolling resistance, so we add drag and Magnus forces to the ball in flight and a constant rolling resistance on the ground. Each expert is trained on 4 NVIDIA RTX A6000 GPUs for 13 to 53 hours.
Distilling the eight experts into the 5.3M-parameter student takes about 4.4 hours on one GPU (behavior cloning followed by four DAgger rounds), and training the shared adapter takes 11.2 hours on four GPUs.
Appendix gives the physical parameters, motion
data, network architectures, and training hyperparameters.

\subsection{Baselines}
\label{sec:exp_baselines}

\paragraph{Skill learning.}
We compare our skill-specific experts and distilled student against TokenHSI*~\citep{pan2025tokenhsi} and SkillX*~\citep{ye2026skillx}, two adapted baselines trained jointly from scratch across all eight skills using a unified Transformer policy. Both baselines follow their official network architectures and PPO configurations, using our observations, task rewards, and motion data. However, they exclude our skill-specific training strategies, including curricula, failure-weighted sampling, multi-objective advantages, and discriminator labels. To isolate the contribution of DAgger, we additionally evaluate the student trained with behavioral cloning alone.

\paragraph{Transitions.}
Direct switching activates the target skill without the adapter. Expert + STI fine-tunes the target expert with PPO from
predecessor-generated states, following the skill transition-based initialization
of PhysicsFC~\citep{kim2025physicsfc}. An episode starts from a skill-switching (handover) state with
probability 0.7 and from a native initial state otherwise, and the fine-tuning uses
the same sample budget per transition as our adapter. One fine-tuned dribbling
expert serves all four transitions into dribbling. This is our reimplementation of
the published method, not PhysicsFC's code. To study the exit, we also run our adapter checkpoint with the exit head disabled (Pers.) and with post-hoc cutoffs that close the gate a fixed time after the handover. To evaluate the role of parameter sharing, we additionally compare against a per-transition adapter baseline (Per-trans.), which trains a separate adapter for each skill pair on the same frozen student. We also include full fine-tuning (Full FT), which directly updates the unified student for transition adaptation.

\subsection{Atomic Skills and Distillation}
\label{sec:exp_atomic}

Each skill is evaluated on 1,024 cases drawn from the initial states and commands of its expert's training, and all policies run the same cases. Off-ball movement succeeds without a failure and with episode-average velocity and facing errors
below 0.75\,m/s and $35^\circ$. Dribbling has no completion event, so we report its
velocity error and failure rate. Shooting and passing succeed when the launch
azimuth is within $10^\circ$ of the command and the launch speed reaches at least
half the commanded speed; shooting also requires the commanded foot. Heading
requires the head to make the first legal touch and an azimuth error below
$25^\circ$. Trapping uses its success bit, stop-and-turn the moment the ball is
stopped at the foot,
and get-up the recovery flag. The student runs with the adapter off, since no transition precedes these episodes.

\paragraph{Experts.}
Table~\ref{tab:atomic_skills} shows that our skill-specific experts substantially outperform the jointly trained baselines on contact-sensitive tasks such as shooting, passing, and trapping.
However, heading remains challenging even for its specialized expert, suggesting that task-specific training alone does not resolve all difficult ball interactions.
The dribbling results also reveal a trade-off: TokenHSI*
has fewer failures but substantially larger velocity-tracking errors, indicating that retaining ball possession does not necessarily imply accurate command following.

\paragraph{Distillation.}
Behavioral cloning alone substantially degrades several
skills, while DAgger recovers most of the lost performance.
This highlights the importance of supervising the student
on states induced by its own actions, rather than relying
solely on expert demonstrations.
The remaining gaps, particularly in heading, largely reflect
limitations of the original experts rather than distillation.

\begin{figure}[t]
    \centering
    \setlength{\tabcolsep}{1pt}
    \renewcommand{\arraystretch}{0.8}
    \begin{tabular}{cc}

    \includegraphics[width=0.48\linewidth]{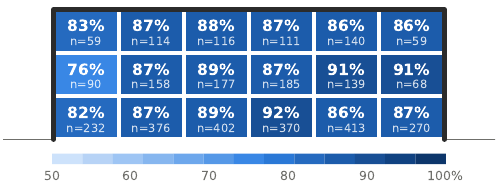} &
    \includegraphics[width=0.48\linewidth]{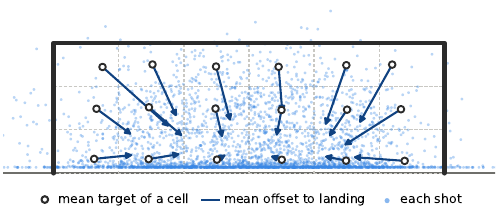} \\[-0.2em]

    (a)  & (b)\\
    \end{tabular}

    \caption{\textbf{Shot placement of our shooting policy}, aiming at a goal 11\,m away (penalty kick).
  (a) Success per commanded target cell ($n$ = shots aimed there). (b) Landing points and the
  mean offset from target to landing per cell.}
      \label{fig:shoot_goal_accuracy}
\end{figure}

\paragraph{Shooting accuracy.}
The shooting bearing metric in Table~\ref{tab:atomic_skills} evaluates the horizontal launch direction, minimum launch speed, and correct kicking foot, but does not assess the ball's final placement. To further evaluate shooting accuracy, we measure where the ball crosses the goal plane relative to the commanded target. We command our policy to shoot and aim at points on a $7.32 \times 2.44$\,m goal 11\,m away, as shown in \Figref{fig:shoot_goal_accuracy}. A
shot succeeds when the ball crosses the goal plane inside the frame, struck with the
commanded foot at no less than half the commanded speed. Of 4,096 shots, 3,479 aim
inside the frame, and 87.2\% of these succeed. The rate varies little across the
goal mouth, between 76\% and 92\% per target cell. The remaining error is
systematic rather than random. The median launch reaches 89\% of the commanded
speed, so high targets land about 1.0\,m low on average, and shots aimed at the posts
drift about 0.9\,m inward.

\begin{figure*}[t]
\centering
\includegraphics[width=1.0\linewidth]{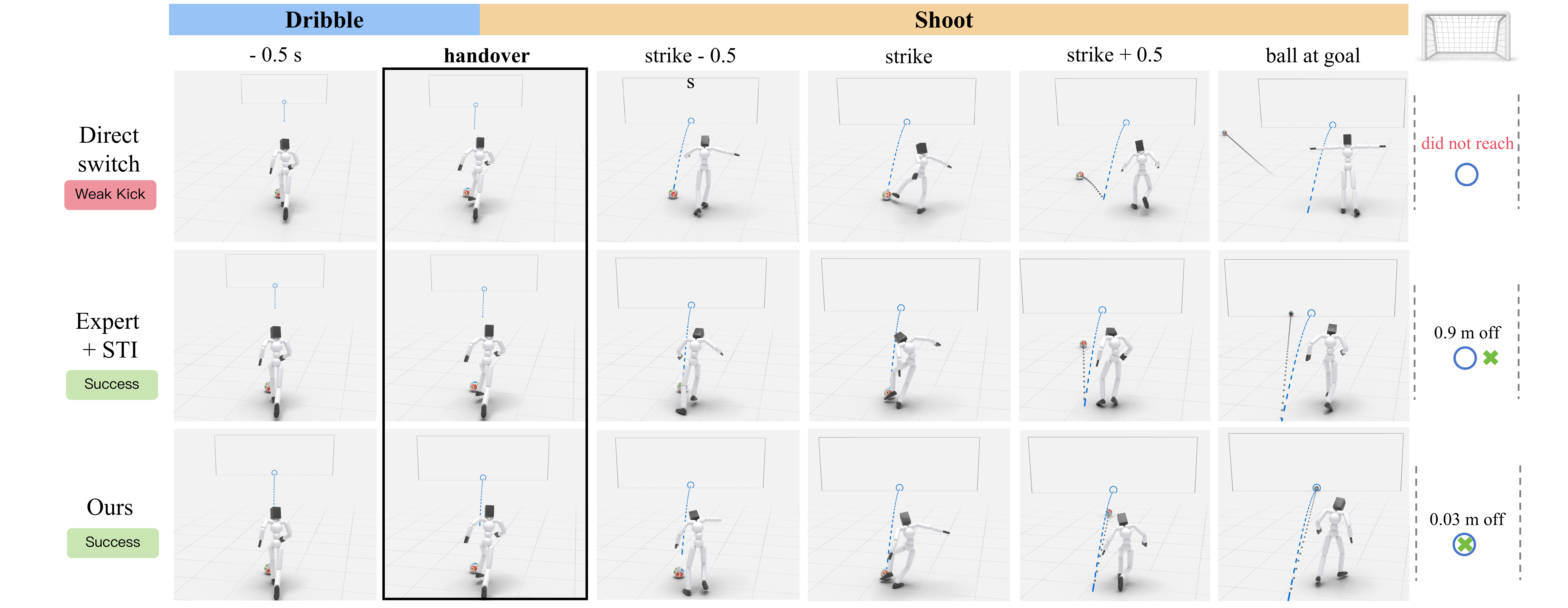}
\caption{\textbf{Dribble$\rightarrow$shoot from the same handover.} The direct switch kicks
weakly; the STI fine-tuned expert scores 0.9\,m from the target, ours 0.03\,m from the target.}
\label{fig:transition_visual}
\end{figure*}


    

\subsection{Skill Transitions}
\label{sec:exp_transitions}

We evaluate five transitions from handover states produced by the frozen student executing the predecessor until its handover condition holds, about 400 to 510 per transition. Every method takes over from the identical states. Dribble$\rightarrow$Shoot succeeds under the native
shooting rule above. A transition into dribbling succeeds when the character has
not fallen and has settled into dribbling by a 4\,s deadline: a velocity error below
0.75\,m/s over the last second, with the ball within 2.5\,m. 
We additionally evaluate each deployed policy under native skill initialization to examine whether transition adaptation affects the original skills. For \paper, the adapter gate remains closed in these episodes, so its actions are identical to those of the distilled student by construction. This evaluation verifies native-mode policy preservation, while the ability of the frozen student to continue execution after an adapter exit is assessed through transition rollouts.

\paragraph{Quantitative results.}

Table~\ref{tab:transitions} highlights the gap between
individual skill competence and transition robustness.
Although the distilled student succeeds in 77.9\% of native
shooting episodes, direct switching from dribbling reduces
success to 19.8\%, exposing a substantial mismatch in
entry-state distributions.
Full fine-tuning improves transition performance but
degrades native shooting success from 77.93\% to 54.82\%,
illustrating the trade-off between adaptation and preserving
previously learned behavior.
Our gated adapter avoids this trade-off in native execution
by keeping the original student unchanged.
Compared with per-transition adaptation, our shared adapter
achieves the highest success rate on four of five transitions,
although specialization benefits dribble$\rightarrow$shoot.
These results suggest that sharing adaptation parameters
can provide competitive performance without separately modifying each skill policy.

\paragraph{Qualitative comparison.}
Since TokenHSI* and SkillX* exhibit poor performance even under native skill initialization, we focus our qualitative comparison on three major transition strategies applied to our trained skills. \Figref{fig:transition_visual} illustrates a dribbling-to-shooting transition under three approaches. Direct switching produces a weak kick that fails to reach the goal. The STI-fine-tuned expert successfully scores but misses the commanded target by 0.9 m. In contrast, our shared adapter achieves a more accurate shot, with the ball reaching within 0.03 m of the target.

\input{tables_arxiv/handoff_ablation}

\subsection{Learned Exit Timing}
\label{sec:exp_exit}
We evaluate our adapter on all 11 transitions under different exit rules, from the same skill switch states. Besides the learned exit and the direct switch, the persistent (Pers.) variant never exits, and post-hoc cutoffs close the gate 0.25 to 2\,s after the skill switch regardless of the exit head.
Each transition has a 4\,s deadline, and kicks are judged 1\,s after the launch. For
stop-and-turn, success is the trap latch of Table~\ref{tab:atomic_skills}. The last
two columns report how often the learned exit returns control before the deadline,
and how long the adapter controls the character on average.

\paragraph{Quantitative results.}

Table~\ref{tab:handoff_ablation_all} reveals that different
target skills require substantially different adaptation
durations. Transitions into off-ball movement typically allow the frozen student to resume control within 0.7\,s, whereas transitions into dribbling rarely trigger an early exit. This suggests that the need for continued adaptation depends strongly on the target skill.

The learned exit improves some transitions over persistent
adaptation, such as shoot$\rightarrow$off-ball, but does not
consistently increase success across all skill pairs.
Its primary benefit is to adapt the duration of intervention
without requiring a manually selected cutoff for each
transition.

\paragraph{Adaptation efficiency.}
Under the same 369M PPO interaction budget,
our shared adapter uses only 0.39M trainable actor parameters (Param.) and 1.5\,MiB of additional storage, about 11$\times$ less than per-transition adaptation.
Despite this reduction, \paper achieves competitive transition performance and higher completion rates across all evaluated skill chains.
Training costs remain comparable (42.7--44.7 GPU-hours), indicating that the primary benefit of sharing is parameter and storage efficiency rather than reduced training computation.

\input{tables_arxiv/adaptation_cost}

%
\input{tables_arxiv/continuous_chains}
\subsection{Continuous Skill Chains}
\label{sec:exp_chains}

\begin{figure*}[h]
    \centering
    \includegraphics[width=1.0\linewidth, trim={0 0.2cm 0 0}]{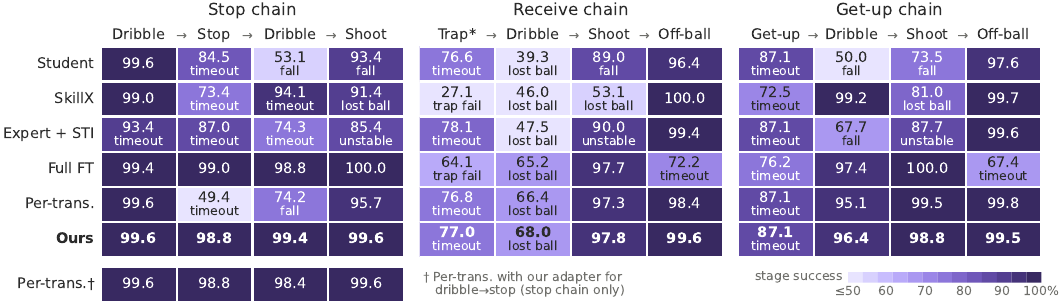}
    \caption{\textbf{Where the chains fail.} Each cell reports the conditional stage success rate (\%), computed over episodes reaching that stage (512 initial episodes per method). The dominant failure type is shown for stages with at least 5\% failures. The hybrid baseline replaces only the dribble-to-stop adapter of Per-trans. with ours, substantially recovering both stop-stage and downstream performance. The receiving stage includes the preceding off-ball approach. Passing chains are omitted for brevity.}
    \label{fig:chain_failures}
\end{figure*}

We evaluate five predefined skill chains in continuous simulation without resetting the character or ball.
A chain succeeds only if all stages meet their completion criteria in sequence: a trap latch for trapping and stop-and-turn, recovery for get-up, 2\,s of dribbling with at least 0.5\,m of progress, ball launch followed by 1\,s of standing for kicking, and velocity and facing tracking for off-ball movement.
All methods use the same criteria and switch skills upon stage completion. This metric measures continuous execution rather than precise kicking
commands, which are evaluated separately in \Secref{sec:exp_atomic} and~\ref{sec:exp_transitions}. Each chain is evaluated over 512 episodes (two seeds). Expert + STI uses the fine-tuned dribbling and shooting experts.

\paragraph{Quantitative results.}
Table~\ref{tab:continuous_chains} shows that strong isolated transitions do not necessarily translate into reliable skill chains. Per-transition adapters achieve only 35.0\% and 35.2\% completion on the two dribble--stop--dribble--kick
chains, compared with 97.5\% and 97.7\% for \paper. Full fine-tuning performs comparably on these chains but degrades on longer chains involving trapping.
These results motivate examining where failures occur during continuous execution.

\paragraph{Failure analysis.}
\Figref{fig:chain_failures} localizes failures by stage. Per-transition adaptation succeeds in only 49.4\% of stop stages, followed by 74.2\% success in subsequent dribbling. Replacing only its dribble-to-stop adapter with ours raises these rates to 98.8\% and 98.4\%, nearly recovering MESSI's performance. This identifies one transition as a major bottleneck in the chain. For reception chains, \paper's failures concentrate in trapping (77.0\%) and subsequent dribbling (68.0\%), while later shooting and off-ball stages exceed 97\%.
Full fine-tuning also struggles with off-ball continuation after shooting (67.4\% \textit{vs.} our 99.5\% on the get-up chain). These findings reveal how failures at individual stages limit long-horizon skill composition.

%% file: tables_arxiv/handoff_ablation.tex
\begin{table}[t]
\centering
\caption{\textbf{Exit rules.} Task success (\%) of one adapter checkpoint on all 11
transitions from identical handover states. Cutoffs close the gate a fixed time after
the handover. Exit: handovers in which the learned exit returns control before the
deadline (\%); On: mean time the adapter is active (s). Skill tags are:
\Mv\ off-ball, \Dr\ dribbling, \Sh\ shooting, \Pa\ passing, \Hd\ heading, \Tr\ trapping,
\ST\ stop-and-turn, \Gu\ get-up. Green / orange: best / second best per row.}
\vspace{-0.5em}
\label{tab:handoff_ablation_all}
\footnotesize
\setlength{\tabcolsep}{2pt}
\resizebox{0.6\linewidth}{!}{%
\begin{tabular}{@{}l*{9}{c}@{}}
\toprule
 & \multicolumn{7}{c}{Task success (\%) $\uparrow$} & \multicolumn{2}{c}{Learned exit} \\
\cmidrule(lr){2-8}\cmidrule(l){9-10}
 & & \multicolumn{4}{c}{Cutoff (s)} & & & & \\
\cmidrule(lr){3-6}
 & Direct & 0.25 & 0.5 & 1 & 2 & Pers. & Ours & Exit & On \\
\midrule
\Dr\nxt\Sh & 19.8 & 26.9 & 45.5 & 61.0 & \best{62.0} & \second{61.8} & \second{61.8} & 10 & 1.61 \\
\Dr\nxt\Pa & 22.2 & 28.0 & 37.1 & \second{37.6} & \best{38.0} & \best{38.0} & \best{38.0} & 1 & 1.46 \\
\addlinespace[2pt]
\Dr\nxt\ST & 82.4 & 95.7 & 96.3 & 97.1 & \second{97.8} & 97.6 & \best{99.2} & 100 & 0.96 \\
\addlinespace[2pt]
\ST\nxt\Dr & 44.1 & 48.4 & 69.3 & 88.2 & \second{96.1} & \best{97.0} & \best{97.0} & 1 & 3.98 \\
\Tr\nxt\Dr & 20.7 & 32.4 & 34.5 & 42.2 & \second{47.3} & \best{50.4} & \best{50.4} & 2 & 3.77 \\
\Gu\nxt\Dr & 32.6 & 53.3 & 73.8 & 83.1 & \second{90.5} & \best{91.7} & \best{91.7} & 2 & 3.90 \\
\Mv\nxt\Dr & 62.7 & 78.0 & 84.7 & 91.2 & \second{95.9} & \best{96.3} & \best{96.3} & 1 & 3.98 \\
\addlinespace[2pt]
\Sh\nxt\Mv & 69.5 & 89.6 & 95.2 & 97.8 & \second{98.2} & 94.2 & \best{98.4} & 100 & 0.60 \\
\Pa\nxt\Mv & 70.2 & 89.2 & 93.7 & 95.7 & \best{97.2} & \best{97.2} & \second{96.3} & 100 & 0.66 \\
\Gu\nxt\Mv & 68.3 & 84.8 & 88.2 & 91.1 & \best{92.3} & 90.9 & \second{91.6} & 99 & 0.70 \\
\Hd\nxt\Mv & 61.1 & 74.0 & 81.1 & 83.1 & \second{84.8} & \best{87.5} & 84.1 & 99 & 0.69 \\
\bottomrule
\end{tabular}
}
\end{table}

%% file: tables_arxiv/adaptation_cost.tex
\begin{table}[t]
\centering
\caption{\textbf{Adaptation cost and model footprint.}
All methods use 369M PPO samples.
Parameters count trainable actor weights, including
conditioning and exit modules.
Storage reports adaptation-specific FP32 weights;
GPU-hours cover transition training.}
\vspace{-0.3em}
\label{tab:adaptation_cost}

\small
\setlength{\tabcolsep}{2.5pt}
\renewcommand{\arraystretch}{1.0}
\resizebox{0.95\linewidth}{!}{%
\begin{tabular*}{\columnwidth}{
    @{\extracolsep{\fill}}llrrr@{}
}
\toprule
Method & Modules
& \shortstack{Params\\(M)}
& \shortstack{GPU\\hours}
& \shortstack{Storage\\(MiB)} \\
\midrule
Expert + STI & 5 policies  & 14.28 & 44.1 & 53.5 \\
Full FT      & 1 policy   & 5.34  & 42.7 & 20.4 \\
Per-trans.   & 11 adapters& 4.33  & 43.0 & 16.5 \\
\addlinespace[2pt]
\textbf{Ours}
& \textbf{1 adapter}
& \textbf{0.39}
& 44.7
& \textbf{1.5} \\
\bottomrule
\end{tabular*}
}
\end{table}

%% file: tables_arxiv/continuous_chains.tex
\providecommand{\best}[1]{\cellcolor{green!15}\textbf{#1}}
\providecommand{\second}[1]{\cellcolor{orange!20}#1}

\begin{table}[h]
\centering
\caption{\textbf{Continuous skill-chain completion rates.} Percentage of episodes in which all stages complete sequentially without environment resets (512 episodes per chain). Stage completion is evaluated using skill-specific criteria, while kicking command accuracy is assessed separately. Skill tags are consistent above.}
\vspace{-0.3em}
\label{tab:continuous_chains}
\footnotesize
\setlength{\tabcolsep}{3.pt}
\resizebox{0.85\linewidth}{!}{%
\begin{tabular}{@{}lcccccc@{}}
\toprule
Skill chain  & \shortstack{Student\\ \small{(direct switch)}}  & \shortstack{Expert\\+\,STI}  & Full FT  & Per-trans. & SkillX* & Ours \\
\midrule
\Dr\nxt\ST\nxt\Dr\nxt\Sh          & 41.8 & 51.6      &\second{97.3}  & 35.0 & {62.5} & \best{97.5} \\
\Dr\nxt\ST\nxt\Dr\nxt\Pa         & 38.5 & 58.6    & \second{97.1}  & 35.2   & {63.5} & \best{97.7} \\
\Mv\nxt\Tr\nxt\Dr\nxt\Sh\nxt\Mv  & 25.8 & {33.2}  & 29.5    & \second{48.8}  &  {6.6}  & \best{51.0} \\
\Mv\nxt\Tr\nxt\Dr\nxt\Pa\nxt\Mv   & 23.8 & {34.6} & 28.5   & \second{48.0}  & {7.8}  & \best{50.6} \\
\Gu\nxt\Dr\nxt\Sh\nxt\Mv         & 31.2   & 51.6     & 50.0       & \second{82.4}   & {58.0} & \best{82.6} \\
\bottomrule
\end{tabular}
}
\end{table}

%% file: sec_arxiv/5_conclusion.tex
\section{Conclusion}
\label{sec:conclusion}

We presented \paper, a unified physics-based soccer controller
that integrates eight task-competent skills and adapts their
transitions through a shared, gated adapter.
Our method improves transition robustness while preserving
the original student policy during native execution.
Experiments further show that reliable skill composition
depends on overcoming specific transition bottlenecks,
rather than optimizing individual transitions in isolation.


%% file: sec_arxiv/X_suppl.tex
\clearpage
\setcounter{page}{1}

\begin{center}
    {\LARGE\bfseries Appendix\par}
\end{center}
\vspace{1em}

\appendix
\noindent{\large\bfseries Contents}
\vspace{0.5em}


\appsec{app:implementation}{Implementation Details}
\appsub{app:simulation}{Simulation, Character, and Ball}
\appsub{app:motion}{Motion Data}
\appsub{app:training}{Expert Training}
\appsub{app:distillation}{Student Distillation}
\appsub{app:adapter}{Transition Adapter}
\appsub{app:hyperparameters}{Hyperparameters and Compute}
\appsub{app:baselines}{Baseline Implementations}


\appsec{app:metrics}{Evaluation Metrics}
\appsub{app:eval_atomic}{Atomic skills.}
\appsub{app:eval_transition}{Transitions}
\appsub{app:eval_native}{Native retention.}
\appsub{app:eval_chains}{Continuous Chain Evaluation}

\appsec{app:skills}{Skill-specific Expert Details}
\appsub{app:skill_offball}{Off-Ball Movement}
\appsub{app:skill_dribble}{Dribbling}
\appsub{app:skill_shoot}{Shooting}
\appsub{app:skill_pass}{Passing}
\appsub{app:skill_header}{Heading}
\appsub{app:skill_trap}{Trapping}
\appsub{app:skill_stop}{Stop-and-Turn}
\appsub{app:skill_getup}{Get-Up}


\appsec{app:supp_exp}{Additional Experiments}
\appsub{app:supp_full_transition}{Complete Skill Transition Evaluation}
\appsub{app:supp_vis_transition}{More Visual Results of Skill Transitions}
\appsub{app:exp_ablation_transition}{Ablation of Transition Adaptation}
\appsub{app:slalom}{Slalom Dribbling Analysis}
\appsub{app:exp_skill_chains}{Additional Results of Skill Chains}

\appsec{app:limitations}{Limitations and Future Work}

\vspace{0.5em}
\noindent\rule{\columnwidth}{0.4pt}

\vspace{1em}

\input{sec_arxiv/Supp/X_1_implementation}

\input{sec_arxiv/Supp/X_5_eval}

\input{sec_arxiv/Supp/X_2_experts}

\input{sec_arxiv/Supp/X_3_results}
\input{sec_arxiv/Supp/X_4_failure}

%% file: sec_arxiv/Supp/X_1_implementation.tex
\section{Implementation Details}
\label{app:implementation}

\subsection{Simulation, Character, and Ball}
\label{app:simulation}

\paragraph{Simulation.}
All experiments are performed in Isaac Lab~\citep{mittal2023orbit} (Isaac Sim~5.1)
with GPU PhysX, using a codebase built on MimicKit~\citep{peng2025mimickit}. The
simulation runs at 240\,Hz, and all policies are queried at 60\,Hz. The physics solver
uses four position iterations and one velocity iteration, with continuous collision
detection enabled.

\paragraph{Character.}
\begin{wrapfigure}{r}{0.5\textwidth}
  \centering
  \includegraphics[width=1.0\linewidth]{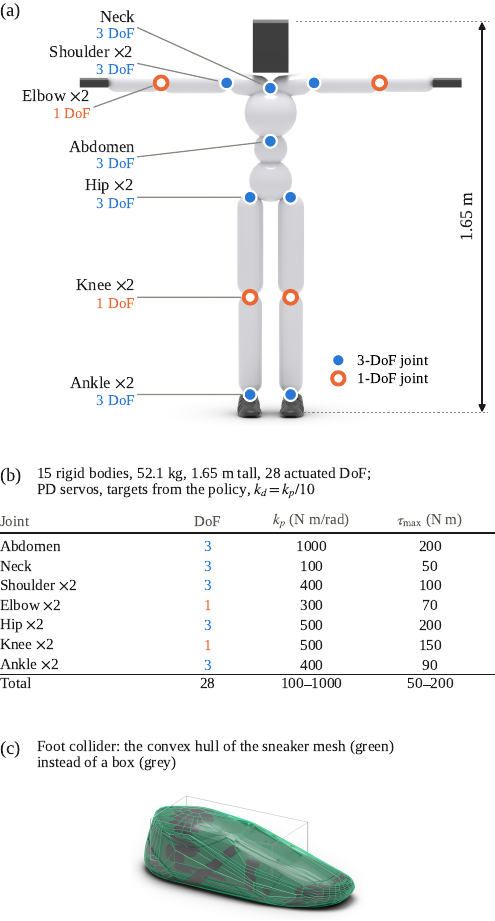}
  \caption{\textbf{Character.} (a)~The zero pose, a T-pose, with the actuated joints and their degrees of freedom.
  (b)~Stiffness $k_p$ and torque limit $\tau_{\max}$ of each joint ($k_d = k_p/10$).
  (c)~The foot collider: the convex hull of the sneaker mesh, which replaces a box.}
  \label{fig:character}
\end{wrapfigure}

The character is a humanoid with 15 rigid bodies, a total mass of 52.1\,kg, and
a standing height of 1.65\,m (Fig.~\ref{fig:character}a). It has 28 actuated
degrees of freedom: the abdomen, neck, shoulders, hips, and ankles have three
each, and the elbows and knees have one each. Every joint is driven by a PD
controller, and the policy specifies the target angles. The stiffness $k_p$
ranges from 100 to 1000\,N\,m/rad, the damping is $k_d = k_p/10$, and the torque
limits range from 50 to 200\,N\,m; Fig.~\ref{fig:character}b lists them per
joint. Instead of boxes, the feet use the convex hulls of sneaker meshes as
collision shapes, so that contacts with the ball follow the shape of the shoe
(Fig.~\ref{fig:character}c).

\paragraph{Ball.}
The ball is a size-5 soccer ball with radius $R = 0.11$\,m and mass 0.45\,kg. Since
the physics engine models neither aerodynamics nor rolling resistance, we apply three
additional forces to the ball at every policy step. Let $\mathbf{v}$ and
$\boldsymbol{\omega}$ denote the linear and angular velocity of the ball, $\rho_a$ the
air density, and $A = \pi R^2$ the cross-sectional area. When the ball moves faster
than 1\,m/s, a quadratic drag force
\begin{equation*}
\mathbf{F}_d = -\tfrac12 \rho_a A C_d \lVert\mathbf{v}\rVert\mathbf{v}
\end{equation*}
is applied. When the ball is also in the air, \ie, its center is more than 3\,cm above
its resting height, a Magnus force
\begin{equation*}
\mathbf{F}_m = \tfrac12 \rho_a A C_l \lVert\mathbf{v}\rVert^2\,
\widehat{\boldsymbol{\omega}\times\mathbf{v}}, \quad
C_l = \min\bigl(C_l^{\max},\, R\lVert\boldsymbol{\omega}\rVert / \lVert\mathbf{v}\rVert\bigr)
\end{equation*}
is applied as well, where the lift coefficient follows wind-tunnel measurements of
soccer balls~\citep{goff2009trajectory}. On the ground, a constant deceleration of
$0.15\,g$ opposes the horizontal velocity until the ball comes to rest. As a
reference, drag shortens a 25\,m/s kick at $16^\circ$ elevation from about 34\,m to
26\,m. We also raise the maximum angular velocity of the ball to
$6\times10^4\,^\circ$/s, so that its spin is not clipped by the simulator.
Table~\ref{tab:impl_physics} summarizes the parameters.

\begin{table}[t]
\centering
\caption{Physical parameters of the ball and the ground. The air density is
$\rho_a = 1.2$\,kg/m$^3$. When two materials are in contact, their friction
coefficients are averaged and the larger restitution is used. The character uses the
default material (friction 1.0, restitution 0), so contacts of the ball with both the
ground and the feet have an effective friction of 0.725 and a restitution of 0.65.}
\label{tab:impl_physics}
\small
\begin{tabular}{lcc}
\toprule
Coefficient & Ball & Ground \\
\midrule
Friction                         & 0.45        & 1.0 \\
Restitution                      & 0.65        & 0.0 \\
Linear damping                   & 0.0         & -- \\
Angular damping                  & 0.05        & -- \\
Rolling resistance               & $0.15\,g$   & -- \\
Drag coefficient $C_d$           & 0.25        & -- \\
Max.\ lift coefficient $C_l^{\max}$ & 0.35     & -- \\
\bottomrule
\end{tabular}
\end{table}

\subsection{Motion Data}
\label{app:motion}

All reference motions are retargeted to our character and played back at 60\,fps.
The only exception is a set of 22 slowed-down LaFAN1 walking clips, which are stored
at 30\,fps and interpolated during playback. The soccer motions are collected from
Soccer-X~\citep{yang2025smgdiff}, two commercial motion capture
datasets~\citep{ultimate_soccer,aa_soccer}, and videos reconstructed with
GVHMR~\citep{shen2024gvhmr}. Foot floating in these clips was corrected manually in
Blender. Some clips also record the trajectory of the ball, which we use to
initialize the ball when an episode starts from a reference state.

The off-ball dataset consists of 127 locomotion clips: 65 from the ACCAD subset of
AMASS~\citep{mahmood2019amass}, 24 from LaFAN1~\citep{harvey2020robust}, 23 from the
example motions of MimicKit, and 15 soccer clips, including mirrored copies. The
trapping expert reuses these clips. The get-up dataset contains 297 clips, 290 of
which are LaFAN1 get-up segments, half of them mirrored. Get-up episodes are
initialized from a separate set of 584 fall clips, which are not shown to the
discriminator. In total, the eight experts use 526 distinct clips (32\,min).
Table~\ref{tab:impl_experts} lists the data used by each skill.

\begin{table}[t]
\centering
\caption{Training setup of the skill experts. Len.\ is the maximum episode length.
Clips and Data refer to the motion data used by the discriminator, and Samples and
Time to the training budget of each expert.}
\label{tab:impl_experts}
\small
\setlength{\tabcolsep}{4pt}
\begin{tabular}{@{}lrrrrr@{}}
\toprule
Skill & Len. & Clips & Data  & Samples & Time \\
      & (s)  &       & (min) & (B)     & (h)  \\
\midrule
Off-ball      & 20.0       & 127 & 14.6 & 5.90 & 21.0 \\
Dribbling     & 20.0$^{a}$ & 26  & 0.9  & 0.73 & 13.2 \\
Shooting      & 10.0       & 10  & 0.5  & 2.12 & 24.4 \\
Passing       & 3.0        & 22  & 0.6  & 4.17 & 52.9 \\
Heading       & 2.5        & 8   & 0.1  & 3.41 & 30.8 \\
Trapping      & 2.2$^{b}$  & 158 & 15.2 & 5.82 & 39.4 \\
Stop-and-turn & 4.0        & 30  & 1.1  & 2.83 & 19.4 \\
Get-up        & 6.0        & 297 & 14.5 & 6.55 & 35.5 \\
\bottomrule
\end{tabular}
\par\smallskip
{\footnotesize\raggedright
$^{a}$\,During training, an episode also ends with its reference clip (at most
4.5\,s).
$^{b}$\,An episode ends 0.5\,s after the planned arrival of the ball.\par}
\end{table}

\subsection{Expert Training}
\label{app:training}

Each expert is trained with PPO~\citep{schulman2017ppo} and an adversarial motion
prior~\citep{peng2021amp}. Following Composite Motion Learning~\citep{xu2023composite},
the task and style rewards are treated as separate objectives. Each objective has its
own value head, and its advantages are normalized separately, so that the objective
weights, rather than the reward scales, determine the relative influence of each
objective. We refer to this multi-objective variant of AMP as MO-AMP
(Table~\ref{tab:impl_hparams}). The task and style objectives are weighted equally. Shooting and passing
add a leg-pose objective, with weights of 0.4, 0.4, and 0.2 for task, style, and pose.
Heading adds a pose objective that also covers the neck, abdomen, and shoulders, with
weights of 0.5, 0.3, and 0.2 (Appendix~\ref{app:skill_header}). The style reward is
$r^{\text{style}} = -\log\max(1-D, 10^{-4})$.

The discriminator observes only the character, not the ball. Its input is a window of
body states expressed relative to the root at the last frame, and it also includes
joint velocities for shooting, trapping, off-ball movement, and get-up. 
For shooting and passing, the discriminator is also conditioned on the kicking technique, and a leg-pose objective encourages consistency with reference motions.
Heading uses a pose objective emphasizing the neck, abdomen, and shoulders. The
trapping discriminator is conditioned on the ball class, and the off-ball
discriminator on whether the character is moving or standing.

The policy outputs normalized actions. For each single-axis joint, the action is
mapped to a PD target around the middle of the joint range, and each three-axis joint
interprets its three actions as an exponential map. The action standard deviation in
Table~\ref{tab:impl_hparams} is given in these normalized units. Observations are
normalized with running statistics, which are frozen after the first $10^8$ samples.
No domain randomization is applied.

Each episode is initialized from a reference state sampled within a skill-specific
window, such as the frames before the strike for the kicking skills. Early
termination is triggered when any body other than the feet touches the ground, except
in get-up, which starts on the ground. Stop-and-turn episodes are also terminated when
the ball is more than 2.5\,m away. The off-ball, passing, and trapping experts were
fine-tuned from earlier checkpoints, and Appendix~\ref{app:skills} describes the final
settings of every skill. Each expert is trained with a single random seed.
Table~\ref{tab:impl_hparams} lists the hyperparameters.

\subsection{Student Distillation}
\label{app:distillation}

The student observes the union of the expert observations (Table~\ref{tab:impl_obs}).
All quantities are expressed in the heading frame of the character, and the
observation blocks of inactive skills are filled with zeros before normalization. Each
block except the skill one-hot is projected to a 256-D token by its own linear layer,
and a learned token-type embedding is added to it. Following
TokenHSI~\citep{pan2025tokenhsi}, the skill one-hot is embedded and added to every
token. The tokens are then processed by a Transformer encoder, whose attention is
masked so that every token other than proprioception attends only to itself and to
the proprioception token. The output tokens are concatenated and mapped to the 28-D
action mean by an MLP with one hidden layer of 1024 units.

The student is first trained with behavior cloning on rollouts of each expert in its
own environment. These rollouts use the last curriculum stage of each expert, without
adaptive clip sampling. We then run four rounds of DAgger~\citep{ross2011dagger}. In
each round, the student's mean action is executed, and every visited state is labeled
with the action of the corresponding expert. The data are aggregated across rounds.
After each round, the student is reinitialized and trained from scratch on all
collected data, with a refitted observation normalizer and 5\% of the samples held out
for validation. The loss is the mean squared error between the normalized action means
of the student and the expert.

\begin{table}[t]
\centering
\caption{Observation blocks of the student. Each expert observes proprioception,
the ball, and its own command block. Stop-and-turn has no command.}
\label{tab:impl_obs}
\small
\setlength{\tabcolsep}{4pt}
\begin{tabularx}{\columnwidth}{@{}lr>{\raggedright\arraybackslash}X@{}}
\toprule
Block & Dim. & Contents \\
\midrule
Proprioception & 140 & root height, rotation, and velocities; joint rotations;
                       joint velocities; head, hand, and foot positions \\
Ball           & 7   & presence flag; relative position and velocity \\
Steering       & 5   & direction, speed, and facing (off-ball, dribbling); dribbling
                       repeats its direction as the facing \\
Shooting       & 9   & launch velocity, foot, technique; strike flag and time since
                       the strike \\
Passing        & 7   & launch velocity, technique; strike flag and time since the strike \\
Get-up         & 2   & recovered flag, elapsed time \\
Heading        & 4   & launch velocity; technique label (a single class) \\
Trapping       & 3   & body part, success flag \\
Skill          & 8   & skill one-hot (student only) \\
\midrule
Total          & 185 & \\
\bottomrule
\end{tabularx}
\end{table}

\subsection{Transition Adapter}
\label{app:adapter}

\paragraph{Context.}
The transition context $\mathbf{c}_t$ is an 80-D vector. After the handover, it contains
the one-hot labels of the source and target skills (16-D), flags for the source phase,
the open gate, and the target phase (3-D), and the time since the handover, divided by
5\,s and clipped at 1 (1-D). It also contains the command block of the target skill,
placed at its position among the 30 command dimensions of the unified observation and
with speeds and launch velocities divided by 10 (30-D), together with a mask that marks
this block (30-D). Before the handover, the target entries hold the source skill and its
command, and the time entry counts from the start of the source.

\paragraph{Architecture.}
The low-rank matrices in \Eqref{eq:cond_lora} have rank $r = 16$, and the LoRA scale
is set to $\alpha = r$, so the update is applied without rescaling. LoRA is applied to
ten linear layers of the student: the proprioception tokenizer, the two feed-forward
layers in each Transformer block, and the readout layer. The attention projections,
the ball and task tokenizers, the skill embedding, and the action head are left
unchanged. Each adapted layer has its own conditioner, an MLP with sizes
$80 \to 64 \to 32$ and SiLU activations. The down-projection matrices are initialized
with Kaiming uniform initialization, while the up-projection matrices and the last
layer of each conditioner are initialized to zero, so the adapted policy is identical
to the student at the start of training. The exit head is an MLP with two hidden layers
of 256 and 128 units, which takes the normalized observation and $\mathbf{c}_t$ as
input. Its output weights start at zero and its output bias at $-7$, so that early exits
are rare at the beginning of training, and exits are only allowed after
$\tau_{\min} = 0.2$\,s. The critic is a separate Transformer with the same architecture
as the student. Its output features are concatenated with $\mathbf{c}_t$ and passed
through a 1024-unit layer with SiLU activation before the value output.

\paragraph{Episodes.}
Table~\ref{tab:impl_edges} lists the 11 transitions used for training, together with
their handover and success conditions. Each episode starts with the source skill in the
scene of its expert, initialized as in its training (Appendix~\ref{app:skills}) at the
final ball-curriculum stage and without adaptive clip sampling, with three exceptions.
Off-ball movement starts in the dribbling scene, with the ball placed at rest 2--4\,m
away within $\pm 60^\circ$ of the heading, and is commanded to run toward it at
2\,m/s. Trapping starts with an off-ball approach toward the meeting point, which lasts
until 0.6\,s before the ball arrives but at most 0.3\,s. For get-up$\to$dribble, a ball
is placed at rest 1--1.5\,m from the body. A dribbling source is commanded to move at
1.5\,m/s along its initial heading. While the gate is closed, the student's mean action
is executed without exploration noise, as at test time. After the handover, the episode
fails if the character falls, if the ball is lost (farther than 4\,m, or 2.5\,m for
stop-and-turn, for 0.25\,s; not checked for off-ball movement or after a strike), or if
the target does not succeed in time (3\,s for off-ball movement, 4\,s for dribbling and
stop-and-turn, and 5\,s until the strike for kicks). The penalty $p_t$ is 1 for a fall or
a lost ball and 0.5 for other failures, such as a timeout or a shot with the wrong foot.
The usage cost is $\kappa = 0.02$ per step while the gate is open. 
Successful episodes are truncated with value bootstrapping rather than treated as terminal.
This avoids penalizing early success when task rewards accumulate over time.

\paragraph{Handover conditions.}
In Table~\ref{tab:impl_edges}, the ball is reachable
if it is within 2.5\,m and below 1.5\,m.
All conditions except off-ball tracking require
an upright standing pose (pelvis above 0.75\,m,
no non-foot ground contacts), checked at the handover
frame for dribbling, off-ball, and heading, and
over the preceding 0.15\,s otherwise.
Get-up already requires 0.3\,s of standing, while
heading hands over within 0.6\,s after contact
unless the character falls.
Evaluation criteria are specified separately
in \Secref{sec:exp_transitions} and~\ref{sec:exp_atomic}.

\paragraph{Optimization.}
Each transition is trained in its own worker with 1{,}024 environments. The 11 workers
share four GPUs and average their gradients at every update. Each transition receives
33.6M samples, for a total of 369M. The remaining hyperparameters are listed in
Table~\ref{tab:impl_hparams_student}.

\begin{table}[t]
\centering
\caption{\textbf{Skill transitions used for adapter training.}
    Handover conditions for source skills and success
    criteria for target skills across all 11 transitions.
    }
\label{tab:impl_edges}
\footnotesize
\setlength{\tabcolsep}{4pt}
\begin{tabularx}{\columnwidth}{@{}l>{\raggedright\arraybackslash}X@{}}
\toprule
\multicolumn{2}{@{}l}{Transitions: off-ball$\to$dribble, dribble$\to$\{shoot, pass, stop\},} \\
\multicolumn{2}{@{}l}{stop$\to$dribble, trap$\to$dribble, get-up$\to$\{dribble, off-ball\},} \\
\multicolumn{2}{@{}l}{\{shoot, pass, header\}$\to$off-ball.} \\
\midrule
Source & Hands over when \\
\midrule
Off-ball      & within 1\,m of a loose ball placed 2--4\,m away \\
Dribbling     & 1\,s after the start, ball reachable \\
Stop-and-turn & ball stopped (trap event, App.~\ref{app:skill_stop}), ball reachable \\
Trapping      & success bit set, ball reachable \\
Get-up        & recovered (into dribbling: also ball reachable) \\
Shoot, pass   & 1\,s after the strike \\
Heading       & 0.3\,s after the header, at the latest 0.6\,s \\
\midrule
Target & Succeeds when \\
\midrule
Dribbling     & after 2\,s and 0.5\,m of progress along the command, ball reachable \\
Shoot, pass   & 1\,s after the strike (shooting: with the commanded foot) \\
Stop-and-turn & ball stopped, ball reachable \\
Off-ball      & velocity (0.3\,s average) within 0.5\,m/s and facing within $30^\circ$ of
                the command for 0.3\,s, at least 0.5\,s after the handover \\
\bottomrule
\end{tabularx}
\end{table}

\subsection{Hyperparameters and Compute}
\label{app:hyperparameters}

Tables~\ref{tab:impl_hparams} and~\ref{tab:impl_hparams_student} list the training
hyperparameters. All policies are trained on NVIDIA RTX A6000 GPUs. Each expert is
trained on four GPUs for 13 to 53 hours (Table~\ref{tab:impl_experts}). Distillation
takes 4.4 hours on a single GPU, excluding evaluation, and the adapter is trained on
four GPUs in 11.2 hours. GPU PhysX simulations are reproducible on a fixed machine but
differ across machines. We therefore run every paired comparison, such as the adapter
versus direct switching, within the same job and from identical initial states.

\subsection{Baseline Implementations}
\label{app:baselines}

\paragraph{Unified multi-skill baselines.}
TokenHSI*~\citep{pan2025tokenhsi} and SkillX*~\citep{ye2026skillx} are trained jointly on all eight skills
using our observations, task rewards, and motion data.
Both use four RTX A6000 GPUs with 8,192 environments
and a combined reward of
$0.5r^{\mathrm{task}}+0.5r^{\mathrm{style}}$,
where $r^{\mathrm{style}}=-2\log(1-D)$.
They do not use our skill-specific curricula,
failure-weighted sampling, multi-objective advantages,
Muon optimizer, or technique-specific discriminator labels.
The TokenHSI used one shared discriminator embedded with 8 skills label one-hot, and SkillX used 8 different discriminators for each skills. Both are evaluated using deterministic actions.

\textbf{TokenHSI*.}
We follow the official implementation~\citep{pan2025tokenhsi},
including its 64-D task tokens, four-layer Transformer,
PPO/AMP settings, and two-stage training.
The first stage jointly trains the eight skills from native
initializations (0.82B samples, 30\,h).
The second stage freezes the shared actor and trains
task tokenizers and action-head adapters on six fixed
skill chains (0.12B samples, 4\,h).

\begin{wrapfigure}{r}{0.5\textwidth}
    \centering
    \includegraphics[width=1.0\linewidth,trim=0pt 20pt 0pt 0pt,
            clip]{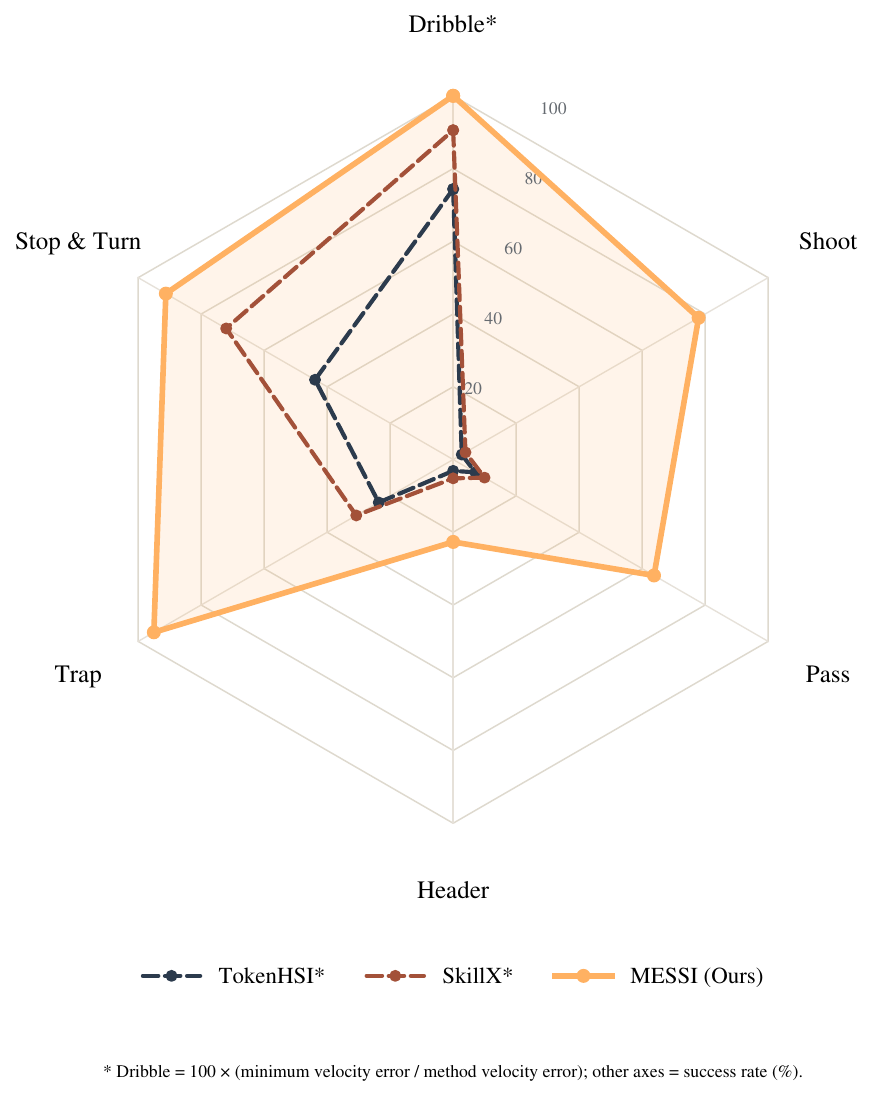}
    
    \caption{\textbf{Atomic skill performance across six soccer skills.}
    Our distilled student outperforms TokenHSI* and SkillX*
    across all six skills under native initialization.
    Dribbling is normalized by inverse velocity error
    ($100\times e_{\min}/e$); other axes show
    skill-specific success rates (\%).
    }
    \label{fig:skill_radar}
\end{wrapfigure}

\textbf{SkillX*.}
As official code is unavailable, we reimplement
SkillX~\citep{ye2026skillx} using its five-frame observation
history, two-layer Transformer, MLP actor, and
KL-adaptive PPO.
Each 20\,s episode switches skills every 5--10\,s
without resetting the character or ball.
To obtain stable training on our tasks, we use reference-state
initialization, replace the discriminator's least-squares
loss with binary cross-entropy, bound the action standard
deviation by 1.0, and terminate failed skills.
We also introduce a training-only ball-loss condition
for dribbling.
Training uses 0.77B samples over 26.1\,h, and we
evaluate the final checkpoint.
The scheduled failure-weighted skill sampling and later
switching curriculum are not reached by this budget.

\paragraph{Transition adaptation baselines.}
Full FT and Per-trans.\ use the same 11 online transitions,
handover conditions, target rewards, and PPO sample
budget as MESSI (33.6M samples per transition).
The frozen student executes the predecessor skill,
and adaptation begins at the handover.

\textbf{Per-trans.}
A separate adapter is trained for each transition,
using the same architecture, exit policy, and
hyperparameters as MESSI.
Each adapter receives 33.6M samples and is deployed
only for its corresponding skill pair.
The hybrid baseline in Fig.~\ref{fig:chain_failures}
replaces only its dribble-to-stop adapter with ours.

\textbf{Full FT.}
We fine-tune all student actor parameters after the
handover, without an adapter or exit policy.
We use a learning rate of $10^{-5}$; a larger rate of
$5\times10^{-5}$ causes transitions into off-ball movement
to collapse.
The fine-tuned policy replaces the original student
throughout evaluation, including native episodes
and continuous skill chains.

\textbf{Expert + STI.}
Following PhysicsFC~\citep{kim2025physicsfc},
we fine-tune five target-skill experts using their
original PPO/AMP configurations.
Training episodes start from predecessor-generated
handover states with probability 0.7 and from native
initializations otherwise.
For target skills with multiple predecessors, a single
expert is jointly fine-tuned on all corresponding
transitions.
Each transition contributes 33.6M PPO samples,
matching the budget of our adapter.
In total, five fine-tuned experts cover all 11 transitions,
using approximately 369M samples.

\begin{table}[t]
\centering
\caption{Hyperparameters of the skill experts, shared by all eight skills unless
noted otherwise. Triples refer to the actor, critic, and discriminator.}
\label{tab:impl_hparams}
\small
\begin{tabular}{@{}ll@{}}
\toprule
Parameter & Value \\
\midrule
Algorithm             & MO-AMP (PPO) \\
Actor and critic      & MLP $4{\times}1024^{a}$ \\
Discriminator         & MLP $2{\times}1024^{b}$ \\
Discriminator window  & 20 frames$^{c}$ \\
Gradient penalty      & 5--20 \\
Logit regularization  & 0.01 \\
Replay buffer         & 200k$^{d}$ \\
Action-rate penalty   & 0--0.1 \\
Action-bound penalty  & 10 \\
Action std.\ (fixed)  & 0.055$^{e}$ \\
Optimizer             & Muon$^{f}$ / Adam / SGD$^{g}$ \\
Learning rate         & $5{\times}10^{-4}$ / $10^{-4}$ / $2.5{\times}10^{-4}$\,$^{h}$ \\
Weight decay          & 0.01 / 0 / $10^{-4}$ \\
Epochs per update     & 5 / 2 / 2 \\
Minibatches per epoch & 8 / 16 / 16 \\
Gradient clipping     & 50 \\
Discount $\gamma$ / GAE $\lambda$~\citep{schulman2016gae} & 0.99 / 0.95 \\
PPO clip ratio        & 0.2 \\
Normalized-advantage clip & 4 \\
Environments          & 8{,}192$^{i}$ \\
Rollout length        & 32 \\
GPUs                  & 4 \\
\bottomrule
\end{tabular}
\par\smallskip
{\footnotesize\raggedright
$^{a}$\,(1024, 512) for off-ball.
$^{b}$\,$2{\times}256$ for passing, heading, and get-up.
$^{c}$\,10 for get-up.
$^{d}$\,100k for heading.
$^{e}$\,0.05 for heading.
$^{f}$\,\citep{jordan2024muon}, applied to weight matrices; biases use AdamW with a
learning rate of $10^{-4}$.
$^{g}$\,Momentum 0.9.
$^{h}$\,$1.25{\times}10^{-4}$ for the discriminators of trapping and stop-and-turn.
$^{i}$\,16{,}384 for off-ball and trapping, 6{,}144 for passing.\par}
\end{table}

\begin{table}[t]
\centering
\caption{Hyperparameters of the student and the adapter. Pairs refer to BC and DAgger
for the student, and to the actor and critic for the adapter.}
\label{tab:impl_hparams_student}
\small
\setlength{\tabcolsep}{4pt}
\begin{tabular}{@{}lll@{}}
\toprule
                      & Student                         & Adapter \\
\midrule
Algorithm             & BC + DAgger$^{a}$               & PPO \\
Network               & Transformer$^{b}$               & LoRA + FiLM$^{c}$ \\
Trainable parameters  & 5.3M                            & 0.39M$^{d}$ \\
Action std.\ (fixed)  & 0.055                           & 0.055 \\
Optimizer             & AdamW$^{e}$                     & Adam \\
Learning rate         & $3{\times}10^{-4}$, cosine$^{f}$ & $3{\times}10^{-4}$ / $10^{-4}$ \\
Discount / GAE        & --                              & 0.99 / 0.995 \\
PPO clip ratio        & --                              & 0.2 \\
Environments          & 512 / 256$^{g}$                 & 1{,}024$^{h}$ \\
Rollout length        & 600 / 400                       & 64 \\
Epochs per update     & 40 / 30                         & 5 / 2 \\
Minibatch size        & 2{,}048                         & 8{,}192 / 4{,}096$^{h}$ \\
Exit head             & --                              & MLP (256, 128) \\
Exit $\tau_{\min}$, cost $\kappa$ & --                    & 0.2\,s, 0.02 \\
Samples               & 2.5M + 4${\times}$0.8M          & 33.6M$^{h}$ \\
GPUs                  & 1                               & 4 \\
\bottomrule
\end{tabular}
\par\smallskip
{\footnotesize\raggedright
$^{a}$\,4 DAgger rounds.
$^{b}$\,4 layers, width 256, 4 heads, feed-forward width 1024.
$^{c}$\,Rank 16 on 10 layers.
$^{d}$\,0.29M in LoRA and the conditioners, 0.10M in the exit head.
$^{e}$\,No weight decay; the gradient norm is clipped at 1.0.
$^{f}$\,Restarted in every round.
$^{g}$\,Per skill.
$^{h}$\,Per transition.\par}
\end{table}

\begin{table}[t]
\centering
\caption{Transition baselines. Full FT and Per-trans.\ train on the 11 transitions of
our adapter; Expert + STI fine-tunes five target-skill experts
covering all 11 transitions. Samples are per transition.}
\label{tab:baselines}
\footnotesize
\setlength{\tabcolsep}{3pt}
\begin{tabularx}{\columnwidth}{@{}l>{\raggedright\arraybackslash}X
                                >{\raggedright\arraybackslash}X r c@{}}
\toprule
Method & Trained & Episodes start from & Samples & Exit \\
\midrule
Ours         & one shared adapter (0.39M)       & online handovers              & 33.6M & learned \\
Per-trans.   & one adapter per transition (11${\times}$0.39M) & online handovers & 33.6M & learned \\
Full FT      & all student actor weights (5.3M) & online handovers             & 33.6M & -- \\
Expert + STI
& five target-skill experts
& 70\% stored handovers, 30\% native
& 33.6M
& -- \\
\bottomrule
\end{tabularx}
\end{table}

%% file: sec_arxiv/Supp/X_5_eval.tex
\section{Evaluation Metrics}
\label{app:metrics}

All policies act with their mean action. Every comparison runs on one machine from
identical initial states (Appendix~\ref{app:hyperparameters}). A \emph{failure} is a
termination by the rules of the skill in control (Appendix~\ref{app:skills}): a fall
for every skill, plus the skill's own failures, such as a ball farther than 2.5\,m in
stop-and-turn. Table~\ref{tab:metric_rules} lists the success rule of every skill.

\subsection{Atomic skills.}
\label{app:eval_atomic}
Each skill is evaluated on 1{,}024 fixed cases drawn from the initial states and
commands of its expert's training. The expert's rollout creates the cases, and every
other policy is accepted only if its initial physical state, task state, and command
sequence match them exactly. An episode runs until the skill's own termination or
horizon (Appendix~\ref{app:skills}). For off-ball movement and dribbling, the
velocity error is the distance between the planar root velocity and the commanded
velocity, and the facing error is the angle between the heading and the commanded
facing. Both are averaged over all frames of the episode, including the transients
after each command change. The failure rate counts episodes that end in a failure.
For the kicks, the launch velocity is the ball velocity at the detected strike
(Appendices~\ref{app:skill_shoot}--\ref{app:skill_header}), and the azimuth error is
the planar angle between the launch and the commanded velocity.

\subsection{Transitions.}
\label{app:eval_transition}
A case is a handover. The frozen student executes the source skill in its own scene
until its handover condition holds (Table~\ref{tab:impl_edges}), and every method
takes over from the identical handover state, which we verify by a hash of the full
state at the handover frame. Two seeds of 256 requests give about 400 to 510
handovers per transition; requests whose source never hands over are not charged to
the target. From the handover, the target runs as a native episode, and its outcome
is fixed at the first failure. A dribbling target is commanded at 1.5\,m/s in a
direction that depends on the source: the reverse of the entry after stop-and-turn,
the initial heading after trapping, the heading at the handover after off-ball
movement, and toward the ball after get-up. An off-ball target is commanded in a uniformly random direction, facing the
direction of travel, at 1--3\,m/s. A kick is commanded along the dribbling direction
at the middle of its technique's ranges, with technique and foot balanced over the
requests. Success is judged within a deadline of 4\,s, and a kick 1\,s after its
launch. For dribbling, the velocity error is averaged over the last second before the
deadline, so the turn of stop$\rightarrow$dribble is not averaged in; for off-ball
movement, it is averaged from the handover, so the rule measures how fast the skill
is established. In Table~\ref{tab:handoff_ablation_all}, \emph{Exit} is the fraction of
handovers in which the learned exit fires before the deadline, and \emph{On} is the
mean time the gate stays open from the handover, up to the deadline. A post-hoc cutoff
closes the gate exactly at its time after the handover.

\subsection{Native retention.}
\label{app:eval_native}
The native columns of Table~\ref{tab:transitions} evaluate each deployed policy on the
cases of Table~\ref{tab:atomic_skills} with the same metrics. A policy that can be
switched off, such as our adapter, runs the student there; a fine-tuned policy runs
itself.


\paragraph{Statistics.}
When two methods are evaluated on the same cases, we compare them with a two-sided
exact sign test. Cases in which both methods succeed or both fail are discarded. If
the methods were equally good, each remaining case would favor either method with
probability 1/2, and we call a difference significant if $p < 0.05$ under this
assumption. The test accounts only for the variation over evaluation cases. Every
trained policy comes from a single training seed, so the test does not cover the
variation between training runs.

\begin{table}[t]
\centering
\caption{Success rules. Native: Table~\ref{tab:atomic_skills}. Transition target:
Tables~\ref{tab:transitions} and~\ref{tab:handoff_ablation_all}, where every rule also
requires no failure before the deadline. Velocity and facing errors are averaged over
the frames given.}
\label{tab:metric_rules}
\footnotesize
\setlength{\tabcolsep}{3pt}
\begin{tabularx}{\columnwidth}{@{}l>{\raggedright\arraybackslash}X
                                >{\raggedright\arraybackslash}X@{}}
\toprule
Skill & Native & As a transition target \\
\midrule
Off-ball      & no failure; velocity error $<0.75$\,m/s and facing error $<35^\circ$ over
                the episode
              & velocity error $<0.75$\,m/s and facing error $<35^\circ$ over the first 4\,s \\
Dribbling     & velocity error and failure rate (no success rule)
              & ball within 2.5\,m throughout; velocity error $<0.75$\,m/s over 3--4\,s \\
Shooting      & azimuth error $<10^\circ$, launch speed $\geq 50\%$ of the command,
                commanded foot
              & as native; no failure until 1\,s after the launch \\
Passing       & azimuth error $<10^\circ$, launch speed $\geq 50\%$ of the command
              & as native; no failure until 1\,s after the launch \\
Heading       & first legal touch detected as a header (App.~\ref{app:skill_header}),
                azimuth error $<25^\circ$ & -- \\
Trapping      & success bit (App.~\ref{app:skill_trap}) & -- \\
Stop-and-turn & trap latch (App.~\ref{app:skill_stop}) & trap latch \\
Get-up        & recovered (App.~\ref{app:skill_getup}) & -- \\
\bottomrule
\end{tabularx}
\end{table}

\subsection{Continuous Chain Evaluation}
\label{app:eval_chains}

Each skill chain runs in continuous simulation for
at most 20\,s, without resetting the character or ball.
The initial skill follows its corresponding expert
environment, while subsequent skills switch upon
stage completion.
Table~\ref{tab:chain_stages} specifies the completion and failure criteria.

The stop chains begin with dribbling, the reception
chains with an off-ball approach to an incoming ball,
and the get-up chain from a fallen state.
Commands for subsequent skills are sampled using
the transition evaluation protocol.

All methods share the same initial states and commands.
Each method executes the first skill using its own policy,
so subsequent handover states may differ.
For MESSI, each skill switch activates the adapter
with a new transition context, and the learned exit
determines when to return control to the frozen student.

\begin{table}[t]
\centering
\caption{Stage tests of the continuous chains. A stage fails on a fall or on the
listed conditions. \emph{Reachable}: the ball is within 2.5\,m and below 1.5\,m;
\emph{standing}: pelvis above 0.75\,m, torso upright, and no body other than the feet
on the ground, here for the last 0.15\,s.}
\label{tab:chain_stages}
\footnotesize
\setlength{\tabcolsep}{3pt}
\begin{tabularx}{\columnwidth}{@{}l>{\raggedright\arraybackslash}X
                                >{\raggedright\arraybackslash}X@{}}
\toprule
Stage & Completes when & Fails when \\
\midrule
\multicolumn{3}{@{}l}{\textit{First stage}} \\
Dribbling     & 1\,s after the start, ball reachable, standing (at that frame)
              & not within 2\,s; ball farther than 4\,m for 0.25\,s \\
Trapping      & success bit set, ball reachable, standing
              & trap failure (App.~\ref{app:skill_trap}); 0.5\,s after the planned arrival \\
Get-up        & recovered, ball reachable & not within 6\,s \\
\midrule
\multicolumn{3}{@{}l}{\textit{Later stages}} \\
Stop-and-turn & ball stopped (trap latch), ball reachable, standing
              & ball farther than 2.5\,m for 0.25\,s; not within 4\,s \\
Dribbling     & after 2\,s and 0.5\,m of progress along the command, ball reachable,
                standing
              & ball farther than 4\,m for 0.25\,s; not within 4\,s \\
Shoot, pass   & 1\,s after the strike, standing (either foot)
              & not standing then; ball farther than 4\,m for 0.25\,s or no strike
                within 5\,s \\
Off-ball      & velocity (0.3\,s average) within 0.5\,m/s and facing within $30^\circ$
                of the command for 0.3\,s, at least 0.5\,s after the handover
              & not within 3\,s \\
\bottomrule
\end{tabularx}
\end{table}

%% file: sec_arxiv/Supp/X_2_experts.tex
\section{Skill Experts}
\label{app:skills}

This section describes the command, reward, initialization, and termination of each
skill expert. Task rewards are weighted sums of terms in $[0, 1]$, except for the
potential-based approach term of trapping, and terms with negative weights are
penalties. We denote the horizontal positions of the root and the ball by
$\mathbf{p}$ and $\mathbf{b}$, their velocities by $\dot{\mathbf{p}}$ and
$\dot{\mathbf{b}}$, and the heading of the character by $\mathbf{h}$. Distances are
in meters, and $[x]_+ = \max(x, 0)$. A fall is a ground contact by any body other than
the feet. Several experts are trained with staged curricula or fine-tuned from earlier
checkpoints, and we describe the final stage of each. Objective weights,
discriminator settings, and PPO hyperparameters are given in
Appendix~\ref{app:implementation}.

\subsection{Off-Ball Movement}
\label{app:skill_offball}

\paragraph{Command.}
The command consists of a travel direction $\mathbf{u}$, a speed $v^*$, and a facing
direction $\mathbf{w}$, all expressed in the heading frame. A new command is sampled
every 4--12\,s. The directions $\mathbf{u}$ and $\mathbf{w}$ are drawn uniformly and
independently, and $v^*$ is drawn uniformly between 0.5\,m/s and a maximum speed that
depends on the angle between $\mathbf{u}$ and $\mathbf{w}$. The maximum is 5.0, 3.5,
2.5, 2.3, and 3.5\,m/s at $0^\circ$, $45^\circ$, $90^\circ$, $135^\circ$, and
$180^\circ$, and is linearly interpolated in between. With probability 0.1, the
character is instead commanded to stand still ($v^* = 0$). After a command that is
more than $45^\circ$ off the facing direction, the next command is a stop with the same
facing with probability 0.2.

\paragraph{Reward.}
Let $\bar{\mathbf{v}}$ be the root velocity $\dot{\mathbf{p}}$ averaged with an
exponential window of 0.3\,s, and let $c = [\mathbf{h}\cdot\mathbf{w}]_+$ measure the
facing alignment. The task reward is
\begin{equation}
r^{\text{task}} = 0.7\,r^{\text{vel}} + 0.3\,r^{\text{face}} + 0.3\,r^{\text{smooth}}.
\end{equation}
For a moving command, the velocity term is
\begin{equation}
\begin{split}
r^{\text{vel}} &= \exp\bigl(-0.5\,\lVert v^*\mathbf{u} - \bar{\mathbf{v}}\rVert^2/\sigma^2\bigr) \\
 &\quad \cdot \bigl(f + (1-f)\,c\bigr)\,g_{\text{step}},
\end{split}
\end{equation}
where $\sigma = \min(0.5\,v^*, 1)$\,m/s, and $r^{\text{vel}} = 0$ if the character
moves backward relative to the command ($\bar{\mathbf{v}}\cdot\mathbf{u} < 0$).
Scaling $\sigma$ with $v^*$ prevents slow commands from being satisfied by standing
still: at $v^* = 0.5$\,m/s, standing earns $e^{-2}$, compared with $e^{-0.125}$ for a
fixed $\sigma = 1$. The facing floor $f$ is 0.2. For lateral commands, which are
$45^\circ$ to $135^\circ$ off the facing direction, $f$ increases linearly to 1 as
$v^*$ goes from 1 to 2\,m/s. The step gate
$g_{\text{step}} = \min(1, \Delta z / 0.03\,\text{m})$ withholds the reward when the
character shuffles without lifting its feet, where $\Delta z$ is the smaller of the two
feet's height ranges over the last 1\,s. The gate is faded out linearly as $v^*$ goes
from 1 to 1.5\,m/s.

For a stand command,
$r^{\text{vel}} = \exp(-128\,\lVert\bar{\mathbf{v}}\rVert^2)\,(0.2 + 0.8\,c)\,
(0.5 + 0.5\,r^{\text{stand}})$. The stance term $r^{\text{stand}}$ is a product of
Gaussians that keep the abdomen and leg joint angles, the pelvis height, and the pelvis
pitch close to the mean of the standing clips, keep the feet level, and, starting
1.5\,s into the command, keep the load balanced between the two feet. The facing term
is $r^{\text{face}} = c\,g_{\text{step}}$. The smoothness term
$r^{\text{smooth}} = \exp(-k\,\overline{\dddot{q}^{\,2}})$ penalizes the mean squared
jerk of the 14 leg joints, with $k = 2\times10^{-8}/(1 + (v^*/1.5)^2)$. The
discriminator is also conditioned on a moving/standing flag. For the policy, the flag
is given by the command, and for a demonstration window, it is set when the mean root
speed is below 0.1\,m/s.

\paragraph{Initialization.}
Each episode starts from a reference state at a random time of one of 127 clips of
running, walking, side-stepping, backpedaling, turning, and standing, including
mirrored copies. Each clip is sampled with a probability proportional to its dataset
weight multiplied by a difficulty score, which is computed from its recent fall rate
and its deviation from the reference. The demonstration windows of the discriminator
are drawn from the same distribution. The ball is placed at rest 10\,m away, and its
observation is set to zero.

\paragraph{Termination.}
An episode ends when the character falls or after 20\,s.

\subsection{Dribbling}
\label{app:skill_dribble}

\paragraph{Command.}
The command consists of a target direction $\mathbf{u}$ and a speed
$v^* \in [0.5, 5]$\,m/s, which are resampled every 4--7\,s. There is no facing
command, and the facing direction is not rewarded.

\paragraph{Reward.}
Let $\mathbf{b}^* = \mathbf{p} + 0.45\,\mathbf{u}$ be the target ball position,
0.45\,m ahead of the character along the command. The error
$\mathbf{e} = \mathbf{b} - \mathbf{b}^*$ is split into a longitudinal component
$e_\parallel = \mathbf{e}\cdot\mathbf{u}$ and a lateral component $e_\perp$. The task
reward is
\begin{align}
r^{\text{lead}} &= \exp\bigl(-(e_\parallel^2 + 4\,e_\perp^2)\bigr), \\
r^{\text{vel}} &= \exp\bigl(-0.5\,\lVert v^*\mathbf{u} - \dot{\mathbf{p}}\rVert^2\bigr), \\
g_{\text{ball}} &= \exp\bigl(-1.5\,[\lVert\mathbf{b} - \mathbf{p}\rVert - 0.6]_+\bigr), \\
r^{\text{task}} &= g_{\text{ball}}\, r^{\text{vel}} + 0.6\, r^{\text{lead}},
\end{align}
where $r^{\text{vel}} = 0$ if $\dot{\mathbf{p}}\cdot\mathbf{u} < 0$. The gate
$g_{\text{ball}}$ equals 1 while the ball is within 0.6\,m of the root, so it does not
penalize the ball distance of normal dribbling. Beyond that distance it decays, to
0.22 at 1.6\,m and 0.05 at 2.6\,m.

\paragraph{Initialization.}
Each episode starts from a reference state at a random time of a dribbling clip. The
ball position is taken from the clip, and its velocity is fitted to 10 frames of the
clip's ball trajectory.

\paragraph{Termination.}
An episode ends when the character falls or after 20\,s. During training, an episode
also ends with its reference clip (Table~\ref{tab:impl_experts}).

\subsection{Shooting}
\label{app:skill_shoot}

\paragraph{Command.}
The command consists of the target launch velocity $\mathbf{v}^*$ in the heading
frame, the striking foot, and the technique, where the latter two are one-hot vectors.
The foot and the technique are those of the reference clip sampled at reset. Instep
shots use speeds of 22--35\,m/s and elevations of $0$--$20^\circ$, and inside shots use
16--26\,m/s and $0$--$15^\circ$. The azimuth is sampled uniformly within $\pm15^\circ$
for instep shots and $\pm20^\circ$ for inside shots, around the heading of the
character at the strike frame of the reference clip. The policy also observes whether
the ball has been struck and the time since the strike, divided by 1\,s and clipped
at 1.

\paragraph{Reward.}
A strike is detected when the ball speed first exceeds 3\,m/s, and the ball velocity
at that moment is stored as $\mathbf{v}_{\text{launch}}$. Let $t$ be the time since
the strike, and let $\rho_{a,b} = \mathrm{clip}\bigl((t-a)/(b-a), 0, 1\bigr)$ be a ramp
that is zero before the strike. The task reward is
\begin{equation}
\begin{split}
r^{\text{task}} &= 0.3\,r^{\text{reach}} + 0.6\,r^{\text{launch}} + 0.15\,r^{\text{foot}} \\
 &\quad + 0.3\,r^{\text{settle}} + 0.3\,r^{\text{rec}}.
\end{split}
\end{equation}
Before the strike, the reach term $r^{\text{reach}} = \exp(-2\,d^2)$ draws a foot to
the ball, where $d$ is the distance from the nearer foot to the ball. After the strike,
$r^{\text{launch}} = \exp(-0.01\,\lVert\mathbf{v}_{\text{launch}} - \mathbf{v}^*\rVert^2)$
scores the launch velocity, and $r^{\text{foot}} = 1$ if the commanded foot made the
strike. Both terms keep their strike-time values for the rest of the episode. The
settle term
\begin{equation}
\begin{split}
r^{\text{settle}} &= \rho_{0.25,0.5}\,
 \exp\bigl(-[\lVert\dot{\mathbf{p}}\rVert - 1.2]_+^2\bigr) \\
 &\quad \cdot \exp\bigl(-100\,(z - 0.86)^2\bigr)
\end{split}
\end{equation}
asks the root to slow down to at most 1.2\,m/s, with its height $z$ close to 0.86\,m.
The recovery term
\begin{equation}
\begin{split}
r^{\text{rec}} &= \rho_{0.6,1.0}\,
 \exp\bigl(-[(z_f - 0.3)/0.15]_+^2 \\
 &\qquad - [(s_f - 4)/2]_+^2\bigr)
\end{split}
\end{equation}
asks for the feet to come back down, where $z_f$ is the larger foot height and $s_f$
the larger foot speed relative to the root. The pose objective (weight 0.2) is
$r^{\text{pose}} = \exp\bigl(-10\sum_i w_i\delta_i^2/\sum_i w_i\bigr)$, where
$\delta_i$ is the angle between the simulated and reference rotations of joint $i$.
The weight $w_i$ is 3 for the ankles, 2 for the knees and hips, and 1 for the other
joints, so the legs dominate the pose objective.

\paragraph{Initialization.}
Each episode starts from a reference state sampled before the strike frame of a clip.
The ball is displaced by up to $\pm0.4$\,m forward and $\pm0.6$\,m sideways in the
heading frame. The displacement shrinks linearly to zero as the sampled time approaches
the strike frame, so the character always has time to adjust its approach. In 40\% of
episodes, the ball starts at rest. Otherwise, it rolls at up to 3\,m/s, with a lateral
speed of up to 0.9\,m/s, in a direction within $\pm90^\circ$ of the motion of the
reference ball, or of the heading if the reference ball is at rest.

\paragraph{Termination.}
An episode ends when the character falls, when the reference clip ends, or after
10\,s.

\subsection{Passing}
\label{app:skill_pass}

\paragraph{Command.}
The command consists of the target launch velocity $\mathbf{v}^*$ in the heading frame
and the technique as a one-hot vector, which is that of the reference clip sampled at
reset. The striking foot is not commanded. The dataset contains every clip together
with its mirror, so either foot can strike. Ground passes use speeds of
4.7--13.5\,m/s and elevations of $0$--$8^\circ$, and the elevation is also capped so
that the vertical launch speed stays below 1\,m/s. Lofted passes use 13.5--24\,m/s and
$18$--$35^\circ$. The azimuth is sampled uniformly within $\pm30^\circ$ for ground
passes and $\pm35^\circ$ for lofted passes, around the launch direction of the
reference clip, which for ground passes often lies well off the heading. The strike
flag and time are observed as for shooting.

\paragraph{Reward.}
Strikes are detected as for shooting, with the additional condition that the ball
velocity changes by more than 1.5\,m/s within one control step. The task reward is
\begin{equation}
\begin{split}
r^{\text{task}} &= 0.03\,r^{\text{reach}}
 + 0.2\,\bigl(r^{\text{speed}} + r^{\text{azim}} + r^{\text{elev}}\bigr) \\
 &\quad + 0.3\,r^{\text{land}} + 0.3\,r^{\text{settle}} + 0.3\,r^{\text{rec}},
\end{split}
\end{equation}
where $r^{\text{reach}}$, $r^{\text{settle}}$, and $r^{\text{rec}}$ are defined as for
shooting. The launch is scored separately for each channel, with
$r^{\text{speed}} = \exp(-(\Delta s/3.5\,\text{m/s})^2)$,
$r^{\text{azim}} = \exp(-(\Delta\psi/25^\circ)^2)$, and
$r^{\text{elev}} = \exp(-(\Delta\theta/12^\circ)^2)$, where $\Delta s$, $\Delta\psi$,
and $\Delta\theta$ are the speed, azimuth, and elevation errors of
$\mathbf{v}_{\text{launch}}$ with respect to $\mathbf{v}^*$. The landing term is
$r^{\text{land}} = \tfrac12\sum_{\sigma\in\{3,8\}\,\text{m}}\exp(-e^2/\sigma^2)$, where
$e$ is the distance between the arrival points predicted for
$\mathbf{v}_{\text{launch}}$ and for $\mathbf{v}^*$ by a ball model with gravity, drag,
and rolling resistance. A ground pass arrives where the ball comes to rest, and a
lofted pass at its first touchdown. The four launch terms are zero before the strike
and keep their strike-time values afterward. The pose objective (weight 0.2) is defined
as for shooting.

\paragraph{Initialization.}
Episodes are initialized as for shooting, with three differences. The ball starts at
rest in 33\% of episodes, its lateral roll speed is at most 0.68\,m/s, and lofted clips
receive an additional forward displacement of 0.3--1.0\,m, which shrinks in the same
way.

\paragraph{Termination.}
An episode ends when the character falls, when the reference clip ends, or after 3\,s.

\subsection{Heading}
\label{app:skill_header}

\paragraph{Command.}
The command is the target outgoing ball velocity $\mathbf{v}^*$ in the heading frame.
The speed is sampled uniformly in 5.5--16.5\,m/s and the elevation in $-32^\circ$ to
$-5^\circ$, so every commanded header goes downward. The azimuth is sampled uniformly
within $\pm110^\circ$ around the heading of the character at the contact frame of the
reference clip.

\paragraph{Reward.}
A header is detected when, within one control step, the ball velocity changes by more
than 3\,m/s, the ball speed exceeds 3\,m/s, and the head is the body closest to the
ball, accounting for body size, and within 0.35\,m of it. Contacts with other bodies
therefore do not trigger the header event. The ball velocity at that step is stored as
$\mathbf{v}_{\text{h}}$. The task reward is
\begin{equation}
\begin{split}
r^{\text{task}} &= 0.05\,r^{\text{reach}} + 0.1\,r^{\text{hit}} + 0.2\,r^{\text{speed}} \\
 &\quad + 0.25\,r^{\text{azim}} + 0.15\,r^{\text{elev}} + 0.2\,r^{\text{settle}}.
\end{split}
\end{equation}
Before the header, $r^{\text{reach}} = \exp(-3\,[d - 0.28]_+^2)$, where $d$ is the
distance from the head to the ball. After the header, $r^{\text{hit}} = 1$, and the
outgoing velocity $\mathbf{v}_{\text{h}}$ is scored separately for each channel as for
passing, with tolerances of 4\,m/s, $40^\circ$, and $20^\circ$ for $r^{\text{speed}}$,
$r^{\text{azim}}$, and $r^{\text{elev}}$. The settle term is defined as for shooting.
The pose objective (weight 0.2) is defined as for shooting, but with weights of 3, 2,
and 1.5 on the neck, abdomen, and shoulders and 1 elsewhere.

\paragraph{Initialization.}
For each clip, the reference state is sampled at a frame at least 0.3\,s before the
contact in which both feet are below 0.16\,m, or at frame 0 if there is no such frame.
In clips with a jump, the policy therefore has to produce the take-off itself. The
ball follows the clip's own cross. Its velocity is a gravity-aware least-squares fit to
the next 10 reference frames, and its start position is shifted so that a drag-free
ballistic flight reaches the reference contact point at the reference contact time.

\paragraph{Termination.}
An episode ends when the character falls, 0.8\,s after the header, or after 2.5\,s.

\subsection{Trapping}
\label{app:skill_trap}

\paragraph{Command.}
The command specifies the receiving body part, foot or chest, as a one-hot vector, and
the policy also observes a success bit. The body part follows from the serve. A ball
that meets the character at a height of 0.9\,m or above is a chest ball, and its first
touch must be with the torso. A lower ball must first touch a foot or a shin.

\paragraph{Reward.}
Let $t_a$ and $\mathbf{p}_c$ be the planned arrival time and meeting point of the
serve. The task reward is
\begin{equation}
\begin{split}
r^{\text{task}} &= 0.3\,r^{\text{app}} + 0.3\,r^{\text{face}} + 2\,r^{\text{touch}}
 + 0.6\,r^{\text{ctrl}} \\
 &\quad + 0.5\,r^{\text{traj}} + 0.2\,r^{\text{stand}}
 - 0.2\,c^{\text{arm}} - 0.3\,c^{\text{plant}}.
\end{split}
\end{equation}
The approach term $r^{\text{app}} = 10\,(0.99\,\phi_t - \phi_{t-1})$ is potential-based
shaping with $\phi = \exp(-(d/0.5)^2)$, where $d$ is the distance from the nearer foot
to $\mathbf{p}_c$, or, for chest balls, from the torso to a point 0.22\,m beyond
$\mathbf{p}_c$. It is paid until 0.25\,s before $t_a$. The facing term
$r^{\text{face}} = \exp(-([\alpha - \alpha_0]_+/25^\circ)^2)$ is paid from 1\,s before
$t_a$ until the first touch or $t_a + 0.2$\,s, whichever comes first. Here, $\alpha$ is
the angle between the facing direction and the incoming ball, and $\alpha_0$ is
$90^\circ$ for foot balls and $30^\circ$ for chest balls. A legal first touch pays
$r^{\text{touch}} = 1$ once. At 0.25\,s after the touch, the ratio $\eta$ of the
ball's planar speed to its incoming speed and the ball's heading $\psi$ relative to the
character's facing at the touch are measured once. From then on, the control term is
\begin{equation}
\begin{split}
r^{\text{ctrl}} &= \exp\bigl(-([|\psi| - 85^\circ]_+/15^\circ)^2\bigr) \\
 &\quad \cdot \exp\bigl(-([\eta - 0.8]_+/0.2)^2\bigr) \\
 &\quad \cdot \bigl(0.5 + 0.5\exp(-((\eta - \eta^*)/0.2)^2)\bigr),
\end{split}
\end{equation}
where $\eta^*$, the speed ratio kept in the clips, is 0.2 for foot balls and 0.6 for
chest balls. The first factor is set to 1 if the ball is slower than 1\,m/s, and
$r^{\text{ctrl}}$ is scaled by a ramp that reaches 1 at 0.3\,s after the touch. The
success bit is set if $\eta \le 0.8$ and either $|\psi| \le 90^\circ$ or the ball is
slower than 1\,m/s, and it is cleared if the episode fails later.

The trajectory term $r^{\text{traj}}$ tracks the median knee and pelvis-pitch
trajectories of the clips with the same ball class (ground, aerial, or chest), from
0.3\,s before $t_a$ to 0.5\,s after the touch. It also caps the root speed toward the
ball at the median of the clips plus 0.2\,m/s. It is paid in standing starts and, in
clip starts, only for aerial and chest balls. The stance term $r^{\text{stand}}$ pays
for a still, upright stance from 0.5\,s after a chest touch. Since an episode ends
0.5\,s after the planned arrival, this term is rarely active in the final stage. The
penalties $c^{\text{arm}}$ and $c^{\text{plant}}$ apply to a hand more than 0.8\,m (foot
balls) or 0.6\,m (chest balls) above the pelvis, and to a support foot that lifts by
more than 2.5\,cm or slides by more than 16\,cm within 0.5\,s of an aerial foot
reception. The discriminator is conditioned on the ball class, which is low, high, or
no ball nearby.

\paragraph{Initialization.}
Most episodes start from one of 30 receiving clips, between 0.2 and 1.2\,s before the
clip's first ball contact, but not before the start of the clip. The clips are sampled
with difficulty-weighted probabilities as for off-ball movement, and the six chest
clips carry 54\% of the base weight. A serve replaces the ball of the clip and keeps
its class. The ball reaches a random point 0.2--1.0\,m ahead of and up to 0.5\,m beside
the character of the clip (chest balls: 0.1--0.35\,m and 0.2\,m), coming from within
$\pm90^\circ$ ($\pm45^\circ$) of the character's facing. It arrives at the contact time
of the clip, or up to 0.5\,s later for balls that come from $60^\circ$ to $90^\circ$ off
the initial facing. Ground balls roll at 4--12\,m/s. Aerial foot balls arrive at
4--12\,m/s and descend at $10$--$60^\circ$, and chest balls arrive at 7--12\,m/s and
descend at $10$--$45^\circ$. Their arrival height is the contact height of the clip
$\pm0.25$\,m, clamped to 0.3--0.85\,m for aerial foot balls and to 1.0--1.35\,m for
chest balls. The launch state is found by integrating the ball model, with drag and
rolling resistance, backward from the meeting point. In 30\% of episodes, the
character instead starts from one of 14 standing clips, and a foot or chest ball, with
equal probability, arrives 0.5--1.2\,s later from within $\pm90^\circ$. It reaches a
point up to 1.5\,m ahead and 1.0\,m to the side (chest balls: 0.8\,m and 0.3\,m).
Serves are biased toward hard cases. About a third take their bearing from the outer
$10^\circ$ of the range, 20\% of the training resets replay a recent failure, and
another 30\% use narrower ranges. These ranges are the top level of two success-gated
serve curricula, at which the final expert is fine-tuned.

\paragraph{Termination.}
An episode fails when the character falls, when a hand or an arm touches the ball, or
when no legal first touch has happened by $t_a + 0.5$\,s. After a legal first touch,
the episode ends at $t_a + 0.5$\,s as a timeout, and its value is bootstrapped. Losing
the ball does not end an episode.

\subsection{Stop-and-Turn}
\label{app:skill_stop}

\paragraph{Command.}
There is no command. The exit direction is fixed to $\mathbf{u} = -\mathbf{u}_0$, the
reverse of the character's heading $\mathbf{u}_0$ at reset.

\paragraph{Reward.}
The reward has two phases, which are separated by the trap event. Let $d$ be the
distance from the nearer foot to the ball. A check is registered when the ball
decelerates by more than 5\,m/s$^2$ beyond rolling resistance while $d < 0.35$\,m.
After a check, the trap indicator $\tau$ latches to 1 at the first step at which
$\lVert\dot{\mathbf{b}}\rVert < 1$\,m/s, $d < 0.45$\,m, and the ball is at most
0.15\,m ahead of the root along $\mathbf{u}_0$. The task reward is
\begin{equation}
\begin{split}
r^{\text{task}} &= (1-\tau)\bigl(0.5\,r^{\text{trap}} + 0.05\,r^{\text{brake}}\bigr) \\
 &\quad + 0.25\,r^{\text{check}} \\
 &\quad + g_{\text{exit}}\,\bigl(0.4\,r^{\text{exit}} + 0.35\,r^{\text{carry}}
 + 0.15\,r^{\text{face}}\bigr).
\end{split}
\end{equation}
Before the trap, $r^{\text{trap}} = \exp(-0.5\,\lVert\dot{\mathbf{b}}\rVert^2 - 8\,d^2)$
rewards a slow ball at the foot, and
$r^{\text{brake}} = \exp(-[\lVert\dot{\mathbf{p}}\rVert - 0.6]_+^2 - 2\,[d - 0.8]_+)$
rewards slowing down while the ball is still close. At every step,
$r^{\text{check}} = \min(1, [a - 1.47]_+/10)\,\mathbb{1}[d < 0.35]$ rewards the ball
deceleration $a$ beyond the 1.47\,m/s$^2$ supplied by rolling resistance. After the
trap, the gate $g_{\text{exit}}$ rises linearly from 0 to 1 between 0.15 and 0.4\,s,
and the exit terms are $r^{\text{exit}} = \mathrm{clip}(\dot{\mathbf{p}}\cdot\mathbf{u}/1.5, 0, 1)$,
$r^{\text{face}} = [\mathbf{h}\cdot\mathbf{u}]_+$, and
$r^{\text{carry}} = \exp(-2\,e_\parallel^2 - 4\,e_\perp^2)$. Here, $e_\parallel$ and
$e_\perp$ are the components along and across $\mathbf{u}$ of
$\mathbf{b} - (\mathbf{p} + \ell\,\mathbf{u})$, with the lead
$\ell = 0.4 + 0.08\,[\dot{\mathbf{p}}\cdot\mathbf{u}]_+$\,m. Without the gate, the
easiest strategy is to turn immediately and leave the ball behind.

\paragraph{Initialization.}
Reset states are taken from the dribbling clips that reach a smoothed root speed of
2\,m/s, at a uniform time between the first frame where this speed is reached and the
next frame where it falls below 0.6\,m/s. The six stop-and-turn clips are used only as
discriminator data, where they make up 60\% of the samples. The root velocity is set to
$v_0\mathbf{u}_0$ with $v_0 \sim \mathcal{U}[2.5, 4.5]$\,m/s, and the joint velocities
are scaled by $v_0$ divided by the reference speed, clamped to $[0.5, 2.5]$. The ball
starts $(0.12 + 0.12\,v_0)\cdot\mathcal{U}[0.55, 1.55]$\,m ahead and up to 0.25\,m to
the side, rolling at $v_0\cdot\mathcal{U}[0.7, 1.3]$ in a direction within
$\pm25^\circ$ of $\mathbf{u}_0$.

\paragraph{Termination.}
An episode ends when the character falls, when the ball is more than 2.5\,m from the
root, or after 4\,s.

\subsection{Get-Up}
\label{app:skill_getup}

\paragraph{Command.}
There is no command. The policy observes a recovered flag and the elapsed time divided
by 3\,s and clipped at 1, and the ball observation is set to zero.

\paragraph{Reward.}
The character counts as recovered once its pelvis has stayed above 0.75\,m for 0.3\,s,
with the vertical component $u_z$ of the torso's up axis above 0.85 and no body other
than the feet touching the ground. The flag then stays on. The task reward is
\begin{equation}
\begin{split}
r^{\text{task}} &= 0.3\,r^{\text{height}} + 0.2\,r^{\text{up}} + r^{\text{prog}} \\
 &\quad + 0.5\,r^{\text{stand}} - 0.5\,c^{\text{late}} - 0.2\,c^{\text{head}},
\end{split}
\end{equation}
with $r^{\text{height}} = \min(z_{\text{head}}/1.3, 1)$ and $r^{\text{up}} = [u_z]_+$.
Before recovery, $r^{\text{prog}} = \min\bigl(20\,[\min(z, 0.85) - z_{\max}]_+, 1\bigr)$
rewards only new records of the pelvis height $z$ above its episode maximum
$z_{\max}$, so rocking up and down earns nothing. After recovery,
$r^{\text{stand}} = \exp(-2\,\lVert\dot{\mathbf{p}}\rVert^2 - 4\,\dot{z}^2
 - 0.5\,\lVert\boldsymbol{\omega}\rVert^2)$ rewards a still root, where
$\dot{\mathbf{p}}$ is the horizontal root velocity and $\boldsymbol{\omega}$ the root
angular velocity. The penalty $c^{\text{late}}$ is 1 while the character has not
recovered after 3\,s, and $c^{\text{head}}$ is 1 while the head touches the ground. The
discriminator data contain no falls. In episodes that start mid-fall, the style reward
is therefore replaced by its batch mean until bodies other than the feet have touched
the ground for a total of 0.2\,s, or until 1.5\,s have passed.

\paragraph{Initialization.}
Reset states are taken from 584 clips (27.4\,min, including mirrored copies): 290
LaFAN1 falls with a sampling weight of 0.5, 290 LaFAN1 get-ups with a weight of 0.4,
and four soccer get-up clips~\citep{aa_soccer} with a weight of 0.1. A fall clip is
sampled within its last 1.5\,s, and the clip ends at most 1\,s after the head first
drops below 0.6\,m, so the character starts mid-fall with the velocities of the clip.
A get-up clip is sampled within its first 40\%, while the character is still on the
ground. The discriminator uses 297 clips (14.5\,min): LaFAN1 get-ups with a share of
0.75, balanced over prone, supine, side, and kneeling starts, the soccer get-ups with
0.1, two AMASS lie-to-crouch clips with 0.05, and one standing clip from
Soccer-X~\citep{yang2025smgdiff} with 0.1. The clips are trimmed before any hop after
standing up.

\paragraph{Termination.}
Before recovery, a fall does not end the episode. After recovery, it does. An episode
also ends after 6\,s.

\begin{table}[t]
\centering
\caption{Skill commands and task objectives. Every expert also receives the
style reward.}
\label{tab:skills}
\footnotesize
\setlength{\tabcolsep}{4pt}
\begin{tabularx}{\columnwidth}{@{}l
  >{\raggedright\arraybackslash\hsize=0.8\hsize}X
  >{\raggedright\arraybackslash\hsize=1.2\hsize}X@{}}
\toprule
Skill & Command $\mathbf{g}^k$ & Task objective \\
\midrule
Off-ball      & direction, speed, facing   & track the velocity and facing \\
Dribbling     & direction, speed           & track the velocity, keep the ball ahead \\
Shooting      & launch velocity, foot, technique & strike with the commanded velocity and foot \\
Passing       & launch velocity, technique & ground or lofted pass at the commanded velocity \\
Heading       & launch velocity            & head an incoming cross at the commanded velocity \\
Trapping      & body part (foot, chest)    & receive the ball with that part and slow it \\
Stop-and-turn & --                         & stop the ball, turn, and dribble it back \\
Get-up        & --                         & stand up from a fall \\
\bottomrule
\end{tabularx}
\end{table}

%% file: sec_arxiv/Supp/X_3_results.tex
\begin{figure*}[t]
    \centering
    \includegraphics[width=1.0\linewidth]{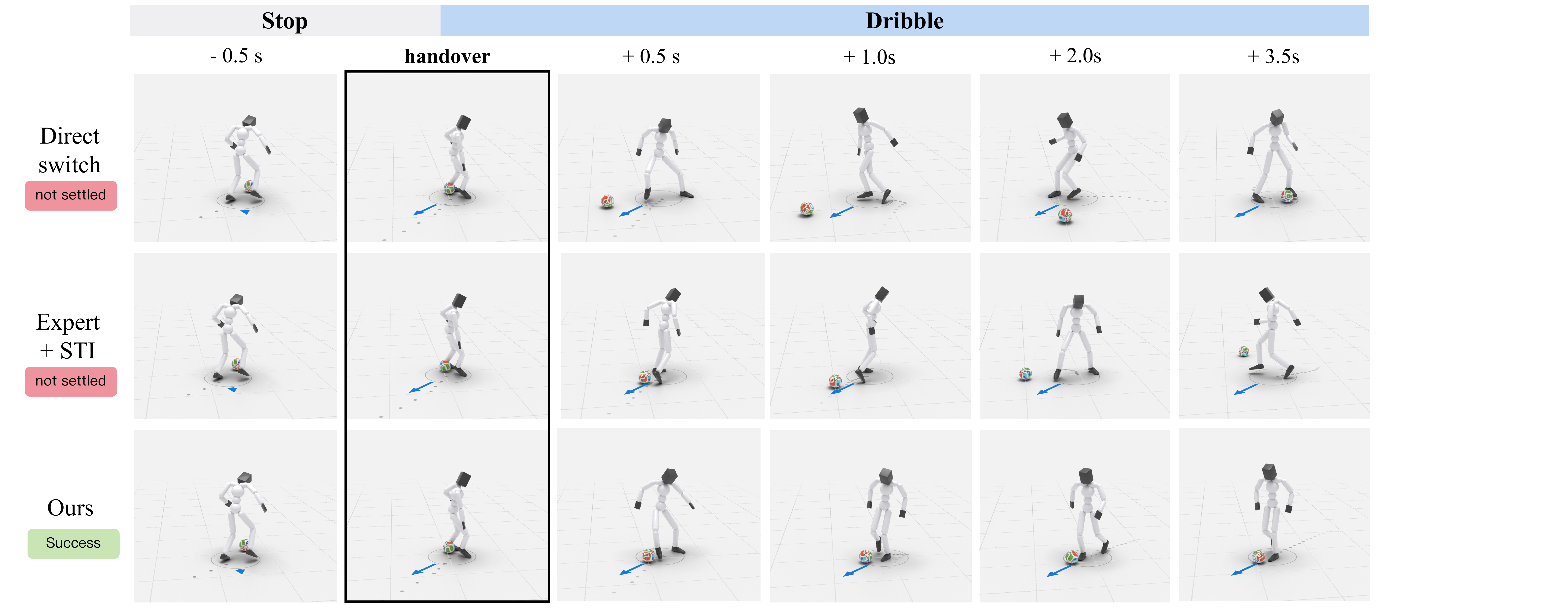}
    \caption{\textbf{Stop$\rightarrow$dribble from the same handover.} 
    }
    \label{fig:transition_02}
\end{figure*}

\section{Additional Experiments}
\label{app:supp_exp}

\input{sec_arxiv/Supp/more_transition}
\subsection{Complete Skill Transition Evaluation}
\label{app:supp_full_transition}
Table~\ref{tab:all_transitions} compares all 11 trained transitions.
MESSI achieves the highest success rate on 9 of
11 transitions, with an average success rate of
82.3\%, compared with 73.1\% for Full FT and
72.5\% for Per-trans.
The exceptions are dribble-to-shoot and
heading-to-off-ball, where Per-trans.\ and SkillX*
perform better, respectively.

\subsection{More Visual Results of Skill Transitions}
\label{app:supp_vis_transition}

\begin{figure*}[h]
    \centering
    \includegraphics[width=1.0\linewidth]{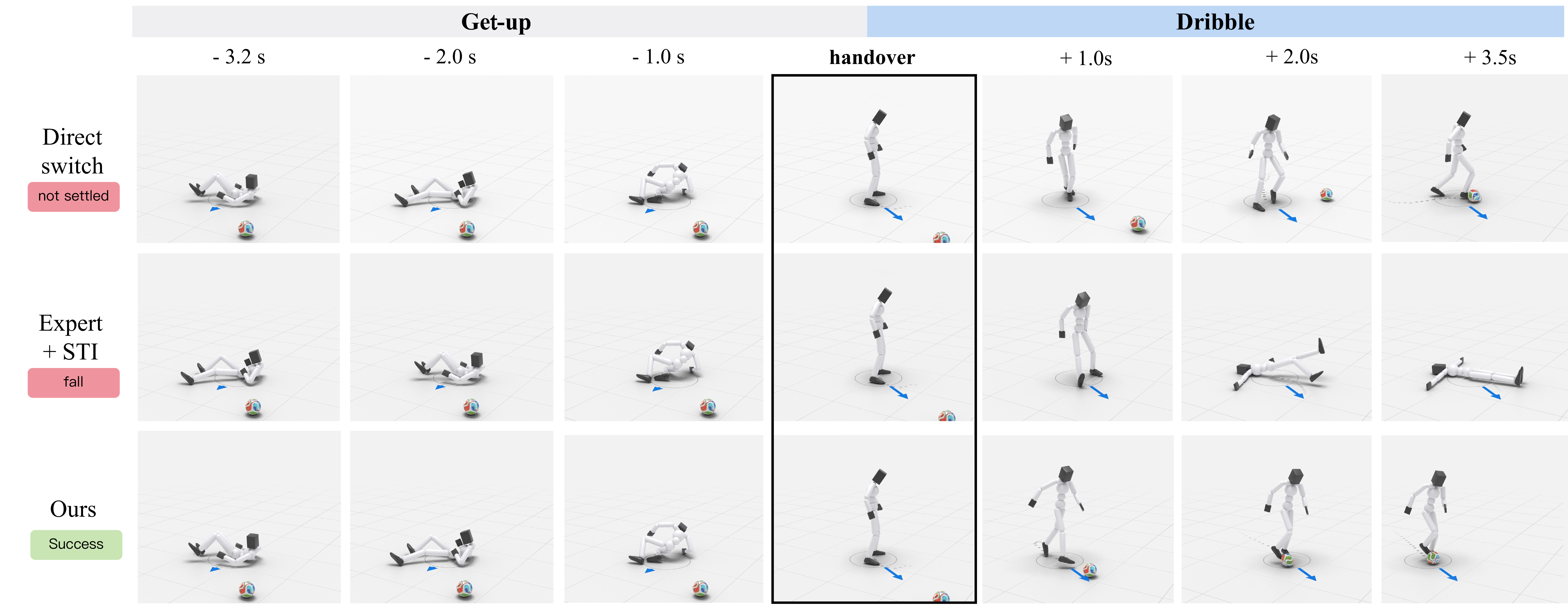}
    \caption{\textbf{Get-up$\rightarrow$dribble from the same handover.} 
    }
    \label{fig:transition_03}
\end{figure*}

Figures~\ref{fig:transition_02} and~\ref{fig:transition_03} provide additional qualitative comparisons of
stop-to-dribble and get-up-to-dribble transitions. All methods are evaluated from identical handover states. For stop-to-dribble, both direct switching and Expert + STI lose ball control during the subsequent dribbling phase, whereas MESSI maintains possession and smoothly resumes dribbling.
For get-up-to-dribble, direct switching fails to establish stable dribbling, while Expert + STI loses balance and falls. In contrast, MESSI successfully transitions from recovery to controlled dribbling. These examples further illustrate the importance of adapting to predecessor-generated states for reliable skill composition.

\begin{table}[h]
\centering
\caption{\textbf{Ablation of transition adaptation.} Task success (\%) on all 11 transitions
from identical handover states. Every variant is one adapter shared by all transitions,
trained with the same budget. Variants differ in whether they are trained with the learned
exit and in the context the adapter's conditioning sees: none, the source and target skills
(pair), or the full context $\mathbf{c}_t$. The exit head always reads $\mathbf{c}_t$.
Green / orange: best / second best per row.}
\label{tab:ctx_ladder}
\footnotesize
\setlength{\tabcolsep}{3pt}
\begin{tabular}{@{}l*{6}{c}@{}}
\toprule
 & & \multicolumn{3}{c}{Without exit} & \multicolumn{2}{c}{With exit} \\
\cmidrule(lr){3-5}\cmidrule(l){6-7}
 & Direct & None & Pair & Full & None & Ours \\
\midrule
\Dr\nxt\Sh & 19.8 & \second{61.2} & 53.1 & 58.4 & 56.9 & \best{61.8} \\
\Dr\nxt\Pa & 22.2 & \second{39.4} & \second{39.4} & \best{41.6} & 35.7 & 38.0 \\
\addlinespace[2pt]
\Dr\nxt\ST & 82.4 & 91.8 & 94.3 & 92.7 & \second{95.7} & \best{99.2} \\
\addlinespace[2pt]
\ST\nxt\Dr & 44.1 & 96.5 & 96.7 & 96.1 & \best{97.3} & \second{97.0} \\
\Tr\nxt\Dr & 20.7 & \best{51.6} & 48.8 & 49.9 & 49.1 & \second{50.4} \\
\Gu\nxt\Dr & 32.6 & 90.5 & \second{91.3} & 90.2 & 91.1 & \best{91.7} \\
\Mv\nxt\Dr & 62.7 & \second{97.4} & \best{97.6} & \second{97.4} & \best{97.6} & 96.3 \\
\addlinespace[2pt]
\Sh\nxt\Mv & 69.5 & 54.3 & 44.1 & 29.6 & \second{97.0} & \best{98.4} \\
\Pa\nxt\Mv & 70.2 & 49.0 & 42.4 & 24.0 & \second{95.6} & \best{96.3} \\
\Gu\nxt\Mv & 68.3 & 44.4 & 41.8 & 25.3 & \second{91.2} & \best{91.6} \\
\Hd\nxt\Mv & 61.1 & 48.1 & 39.0 & 21.9 & \second{83.2} & \best{84.1} \\
\midrule
Average & 50.3 & 65.8 & 62.6 & 57.0 & \second{80.9} & \best{82.3} \\
\bottomrule
\end{tabular}
\end{table}

\subsection{Ablation of Transition Adaptation.}
\label{app:exp_ablation_transition}
Table~\ref{tab:ctx_ladder} evaluates the effects of transition
conditioning and learned exit.
All variants share one adapter across 11 transitions
and use the same training budget.
Conditioning is varied among no context, skill-pair
labels, and the full transition context.

\paragraph{Effect of conditioning.}
Without learned exit, conditioning does not
consistently improve transition success.
The target skill and command are already encoded
in the student's observation, which may reduce
the additional benefit of explicit conditioning.

\paragraph{Effect of learned exit.}
Training with learned exit substantially improves
the average success rate, particularly for transitions
into off-ball movement.
Without exit, persistent adaptation often degrades
locomotion after the initial transition.
The learned exit enables the frozen student to
resume native control.
Since disabling the exit only at test time produces
similar results, the improvement largely arises
from training with the exit mechanism rather than
from its test-time deactivation.

\subsection{Slalom Dribbling Analysis}
\label{app:slalom}

\begin{figure}
    \centering
    \includegraphics[width=0.9\linewidth]{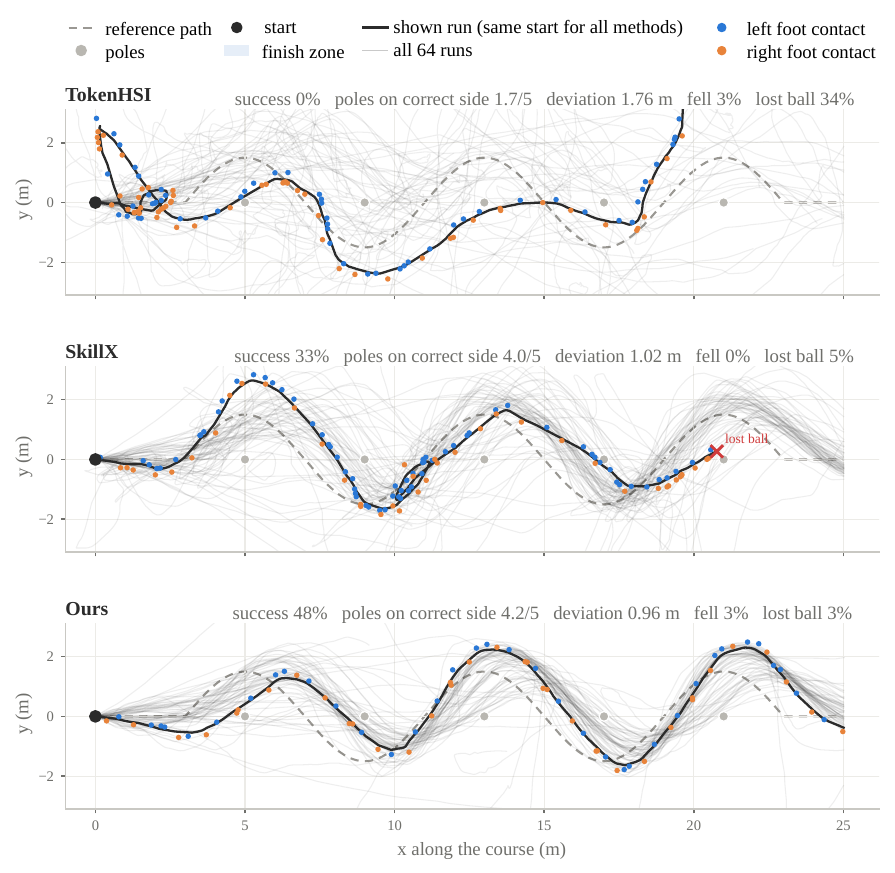}
    \caption{\textbf{Slalom dribbling (top view).} All methods follow the same reference weave (dashed; 5 poles, 4 m apart, ±1.5 m, 1.8 m/s) under the same pure-pursuit steering command, from the same 64 reset states. The shown run starts from the same state for every method; thin lines are all 64 runs; dots mark left / right foot touchdowns. A run succeeds if it finishes without falling or losing the ball and passes every pole on its side. }
    \label{fig:slalom dribbling}
\end{figure}

We evaluate continuous steering and ball control on a slalom course with varying pole spacing.
At 4.0\,m spacing, our student achieves 48\% success, compared with 33\% for SkillX* and 0\% for TokenHSI*. However, SkillX* performs better on tighter courses with 3.0\,m and 2.5\,m spacing (47\%/30\% versus our 5\%/0\%). Our failures are consistent with the heading bias and delayed steering response inherited from the dribbling expert, highlighting a limitation of the underlying skill rather than transition adaptation.

\subsection{Additional Results of Skill Chains}
\label{app:exp_skill_chains}

\input{img/long_chains/chains_figure}
Figure~\ref{fig:long_chain_results} presents four additional
qualitative examples of long-horizon skill composition.
Each sequence involves multiple consecutive soccer skills
executed in continuous simulation without resetting the
character or the ball. The overlaid character poses and
ball trajectories illustrate the coordinated whole-body
motions and ball interactions across successive skill
transitions. These examples complement the quantitative
results in the main paper by visualizing continuous
multi-skill execution.

\paragraph{Strict kick criterion.}
Table~\ref{tab:continuous_chains_strict} repeats Table~\ref{tab:continuous_chains}
with a strict test for the kick. The kick must also use the commanded foot when
shooting, launch within $10^\circ$ of the commanded azimuth, and reach at least half
the commanded speed. Chain success is then bounded by kick accuracy, and every method
completes far fewer chains. Our controller is no longer the best under this criterion.
Full FT leads on the two stop-and-turn chains (85.0\% and 28.7\%, against our 64.8\%
and 18.8\%), and Per-trans.\ leads on the chains that start with trapping or get-up
(26.8\%, 15.6\%, and 55.3\%, against our 19.1\%, 11.3\%, and 35.2\%). Both kick more
accurately after dribbling than our shared adapter. Neither leads under the main
criterion, however: Full FT completes far fewer of the chains with trapping or get-up,
and Per-trans.\ far fewer of the stop-and-turn chains (Table~\ref{tab:continuous_chains}).
SkillX completes many chains under the main criterion but almost none under the
strict one, so its kicks rarely meet the command. The strict rates of our controller
match those of the persistent adapter, so the exit does not cost kick accuracy.

\begin{table}[t]
\centering
\caption{\textbf{Continuous chains, strict kick criterion.} Complete-chain success
(\%), 512 chains per cell, as in Table~\ref{tab:continuous_chains}, but the kick must
also use the commanded foot (shooting), launch within $10^\circ$ of the commanded
azimuth, and reach at least half the commanded speed. Columns are the chains of
Table~\ref{tab:continuous_chains}, read from top to bottom. Green / orange: best /
second best per chain.}
\label{tab:continuous_chains_strict}
\footnotesize
\setlength{\tabcolsep}{4pt}
\begin{tabular}{@{}l*{5}{c}@{\hspace{6pt}}cc@{}}
\toprule
 & \shortstack{\Dr\\\ST\\\Dr\\\Sh} & \shortstack{\Dr\\\ST\\\Dr\\\Pa}
 & \shortstack{\Mv\\\Tr\\\Dr\\\Sh\\\Mv} & \shortstack{\Mv\\\Tr\\\Dr\\\Pa\\\Mv}
 & \shortstack{\Gu\\\Dr\\\Sh\\\Mv} & Avg. & \shortstack{Main\\avg.} \\
\midrule
Expert                          & 6.4           & 12.5          & 4.1           & 5.3           & 3.3           & 6.3           & 32.8 \\
Student                         & 4.9           & 9.2           & 2.9           & 5.9           & 2.9           & 5.2           & 32.2 \\
SkillX~\citep{ye2026skillx}      & 0.0           & 5.5           & 0.2           & 1.2           & 0.0           & 1.4           & 39.6 \\
Expert + STI                    & 21.3          & 16.4          & 10.9          & 9.8           & 18.2          & 15.3          & 45.9 \\
Full FT                         & \best{85.0}   & \best{28.7}   & 17.2          & 8.8           & \second{37.1} & \second{35.4} & 60.5 \\
Per-trans.                      & 21.5          & 8.8           & \best{26.8}   & \best{15.6}   & \best{55.3}   & 25.6          & 49.8 \\
Per-trans.\textsuperscript{\dag} & \second{66.4} & 17.2         & \best{26.8}   & \best{15.6}   & \best{55.3}   & \best{36.3}   & 74.4 \\
Persistent                      & 64.6          & \second{20.3} & 18.9          & \second{11.3} & 35.5          & 30.2          & \second{75.4} \\
Ours                            & 64.8          & 18.8          & \second{19.1} & \second{11.3} & 35.2          & 29.8          & \best{75.9} \\
\bottomrule
\end{tabular}
\par\smallskip
{\footnotesize\raggedright
Avg.: unweighted mean of the five strict rates; Main avg.: the same mean under the main
criterion of Table~\ref{tab:continuous_chains}.
\textsuperscript{\dag}\,Per-trans.\ with its dribble$\to$stop adapter replaced by ours.
Only the first two chains contain that transition, so the last three cells repeat
Per-trans.\par}
\end{table}

%% file: sec_arxiv/Supp/more_transition.tex
\begin{table}[t]
\centering
\caption{\textbf{Complete skill-transition comparison.}
Success rates (\%) across all 11 trained transitions,
evaluated from identical predecessor-generated
handover states.
All adaptation baselines are trained on the same
11-transition set.}
\label{tab:all_transitions}

\small
\setlength{\tabcolsep}{4pt}
\renewcommand{\arraystretch}{1.12}

\begin{tabular}{@{}lcccccc@{}}
\toprule
Transition
& SkillX*
& \shortstack{Direct-\\switch}
& \shortstack{Expert\\+ STI}
& \shortstack{Full\\FT}
& \shortstack{Per-\\trans.}
& \textbf{Ours} \\
\midrule

\Dr\nxt\Sh & 2.7 & 19.8 & 54.1 & 57.3 & \best{67.8} & \second{61.8} \\
\Dr\nxt\Pa & 17.3 & 22.2 &  31.6  & \second{35.7} & 35.1 & \best{38.0} \\
\Dr\nxt\ST & 75.7 & 82.4 & 90.2  & \second{95.9} & 43.5 & \best{99.2} \\

\addlinespace
\ST\nxt\Dr & 95.8 & 44.1 & 63.8 & \second{96.9} & 94.0 & \best{97.0} \\
\Tr\nxt\Dr & 47.9 & 20.7 & 27.7 & \second{49.1} & 47.8 & \best{50.4} \\
\Gu\nxt\Dr & \second{91.3} & 32.6 & 55.7 & 91.0 & 90.8 & \best{91.7} \\
\Mv\nxt\Dr & 83.9  & 62.7 & 59.8 & \second{95.0} & 92.6 & \best{96.3} \\

\addlinespace
\Sh\nxt\Mv & \second{83.4} & 69.5 & 83.1 & 78.0 & 82.5 & \best{98.4} \\
\Pa\nxt\Mv & \second{83.5} & 70.2 & 76.7 & 72.8 & 81.9 & \best{96.3} \\
\Gu\nxt\Mv & 81.4 & 68.3 & 72.5 & 68.3 & \second{86.2} & \best{91.6} \\
\Hd\nxt\Mv & \best{87.4} & 61.1 & 69.8 & 64.5 & 74.9 & \second{84.1} \\ \midrule
\textbf{Average} & 68.2 & 50.3 & 62.3 & \second{73.1} & 72.5 & \best{82.3} \\

\bottomrule
\end{tabular}
\end{table}

%% file: img/long_chains/chains_figure.tex
\captionsetup[subfigure]{skip=3pt}
\captionsetup[figure]{skip=6pt}

\begin{figure*}[p]
    \centering

    \begin{subfigure}{\linewidth}
        \centering
        \includegraphics[
            width=\linewidth,
            trim=0pt 140pt 0pt 20pt,
            clip
        ]{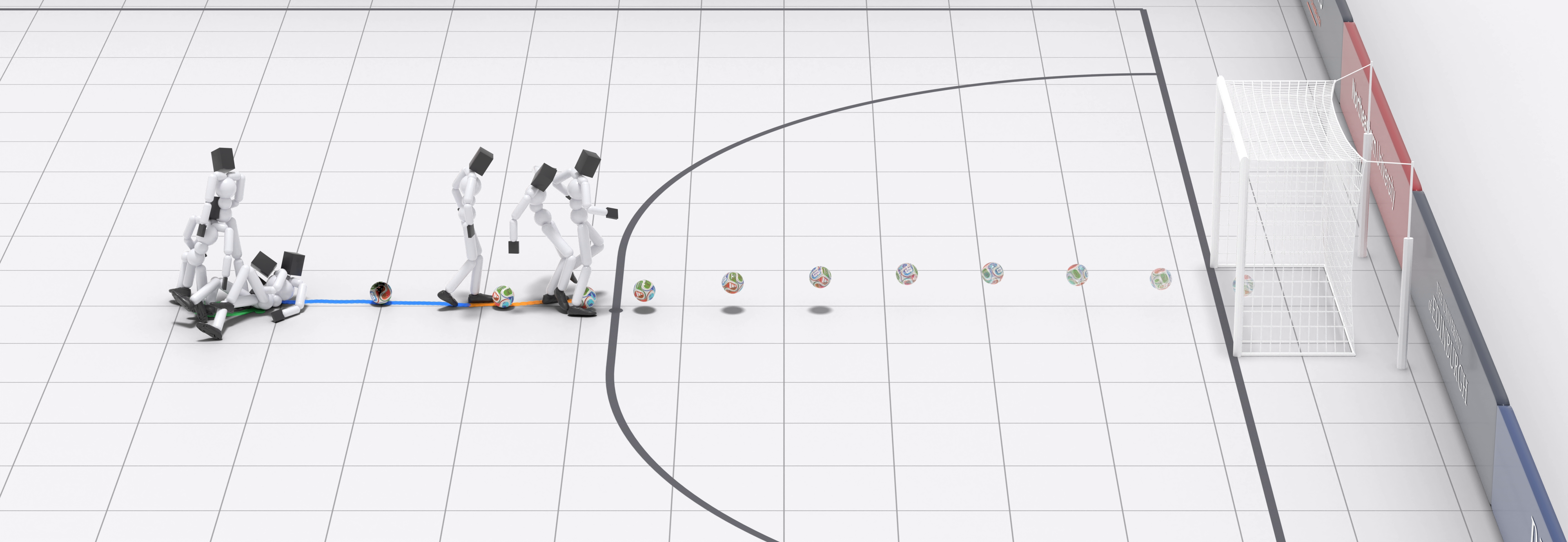}
        \caption{\Gu\nxt \Dr \nxt \Sh}
        \label{fig:chain_a}
    \end{subfigure}

    \par\vspace{0.4em}

    \begin{subfigure}{\linewidth}
        \centering
        \includegraphics[
            width=\linewidth,
            trim=0pt 140pt 0pt 20pt,
            clip
        ]{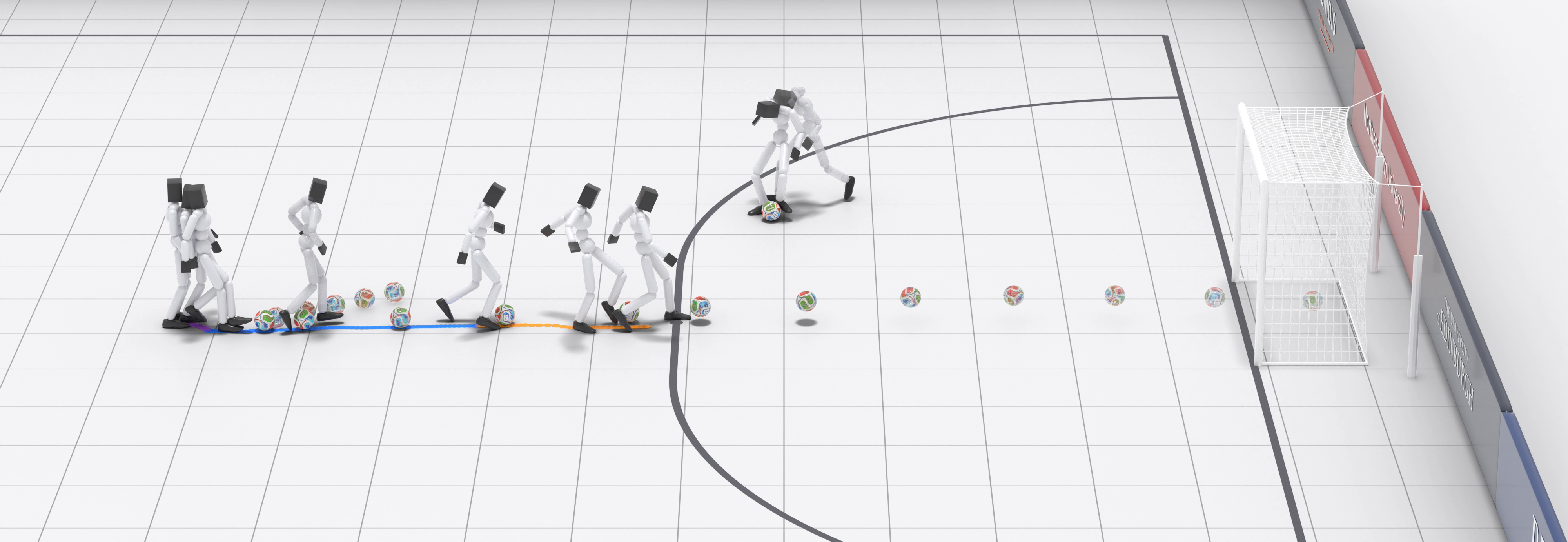}
        \caption{\Tr \nxt \Dr \nxt \Sh}
        \label{fig:chain_b}
    \end{subfigure}

    \par\vspace{0.4em}

    \begin{subfigure}{\linewidth}
        \centering
        \includegraphics[
            width=\linewidth,
            trim=0pt 140pt 0pt 20pt,
            clip
        ]{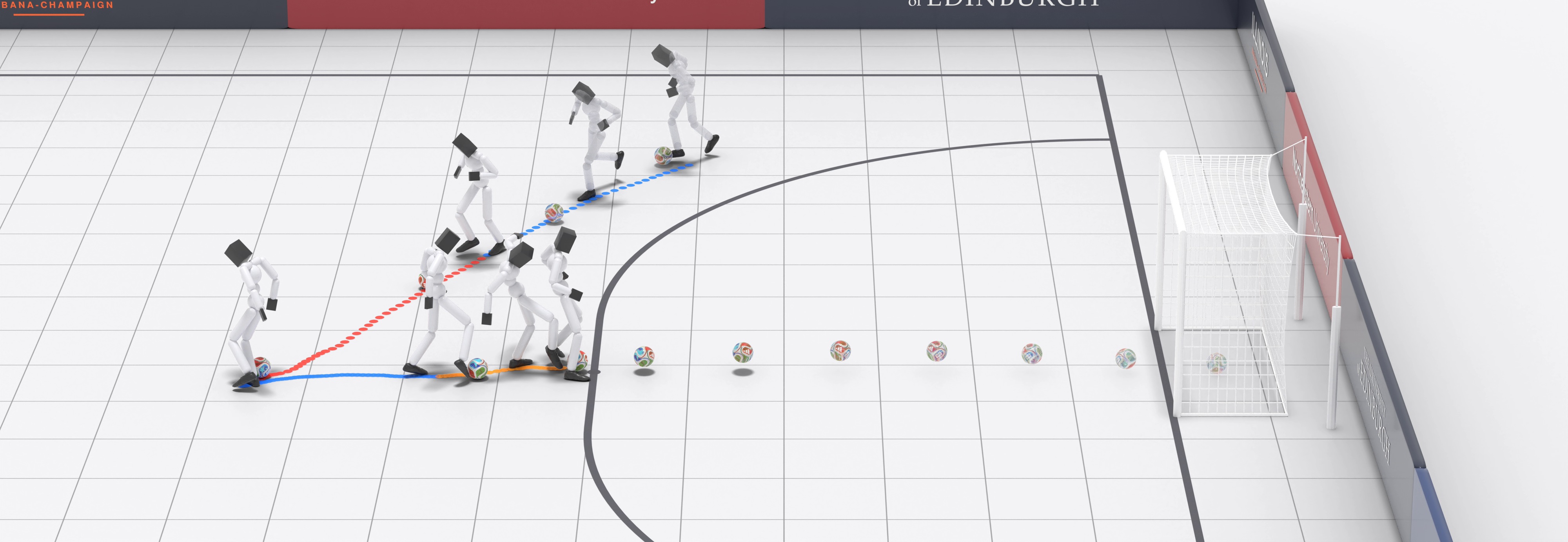}
        \caption{\Dr \nxt \ST \nxt \Dr \nxt \Sh}
        \label{fig:chain_c}
    \end{subfigure}

    \par\vspace{0.4em}

    \begin{subfigure}{\linewidth}
        \centering
        \includegraphics[
            width=\linewidth,
            trim=0pt 140pt 0pt 20pt,
            clip
        ]{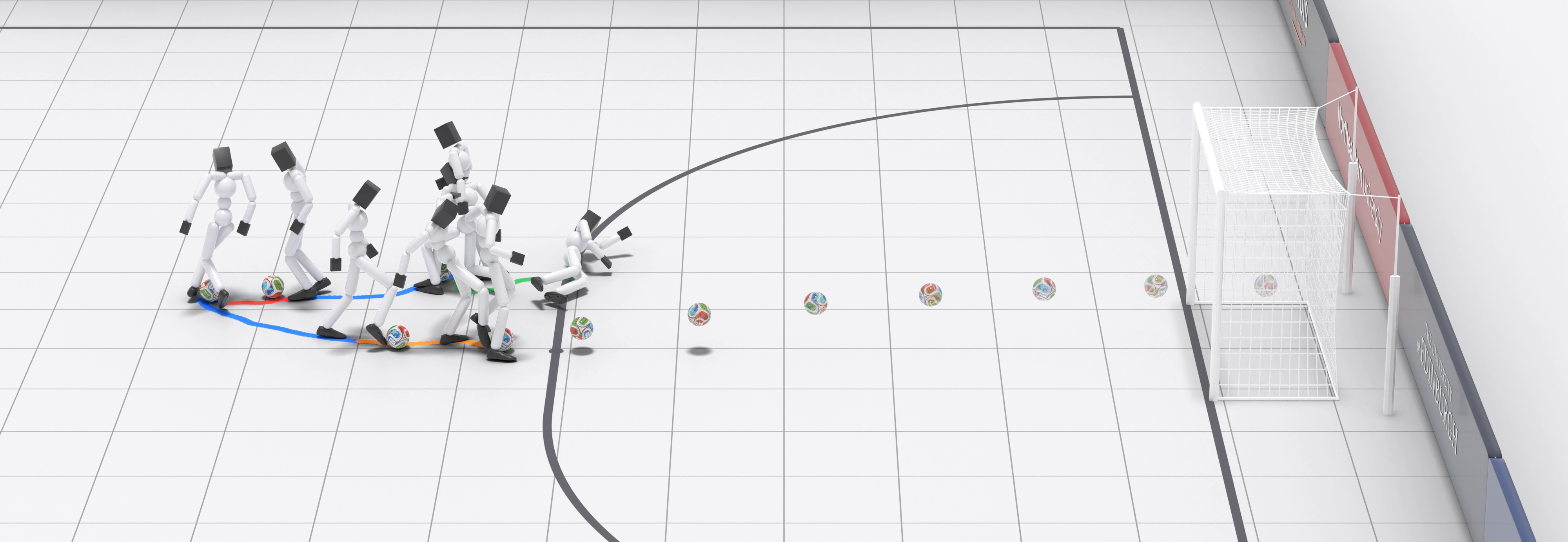}
        \caption{\Gu \nxt \Dr \nxt \ST \nxt \Dr \nxt \Sh}
        \label{fig:chain_d}
    \end{subfigure}

    \caption{\textbf{Additional qualitative results
    of long-horizon skill composition.} Skill tags are: \Dr\ dribbling, \Sh\ shooting,  \Tr\ trapping, \ST\ stop-and-turn, \Gu\ get-up. }
    \label{fig:long_chain_results}
\end{figure*}

%% file: sec_arxiv/Supp/X_4_failure.tex
\section{Limitations and Future Work}
\label{app:limitations}

\paragraph{Limitations.}
MESSI adapts only the 11 trained transitions,
and unseen skill switches may fail.
Several trained transitions remain challenging,
particularly trap-to-dribble and dribble-to-pass.
The student also inherits limitations of its experts,
including weak heading and delayed dribbling
responses.

Our chain-completion metric emphasizes continuous
execution rather than precise kicking commands,
and stop-and-turn is evaluated by ball stopping
rather than the full turning maneuver.
Furthermore, the learned exit rarely deactivates
the adapter during dribbling and kicking.

Training requires skill-specific rewards, curricula,
and manually defined handover conditions.
Adding a new skill requires expert training,
student redistillation, and adapter retraining.
All experiments are conducted in simulation.

\paragraph{Future work.}
Future directions include learning high-level skill
selection, extending adaptation to unseen transitions,
improving rare behaviors, and supporting multiplayer
control and physical humanoids.